%% file: nmg_arxiv.tex
\documentclass[10pt,twocolumn,letterpaper]{article}

\input{include/nmg/packages.tex}

\cvprfinalcopy
\def\cvprPaperID{3053}

\newif\ifarxiv
\arxivtrue

\input{include/nmg/abbrv.tex}

\begin{document}
\title{Multigrid Neural Architectures}
\input{include/nmg/authors.tex}

\maketitle

\begin{abstract}
\input{include/nmg/sec0_abstract.tex}
\end{abstract}

\input{include/nmg/sec1_introduction.tex}
\input{include/nmg/sec2_related.tex}
\input{include/nmg/sec3_multigrid.tex}
\input{include/nmg/sec4_experiments.tex}
\input{include/nmg/sec5_conclusion.tex}

{\small
\bibliographystyle{ieee}
\bibliography{nmg_arxiv}
}

\end{document}

%% file: include/nmg/packages.tex
\usepackage{cvpr}
\usepackage{times}
\usepackage{graphicx}
\usepackage{amsmath}
\usepackage{amssymb}
\usepackage{bm}
\usepackage{makecell}
\usepackage{comment}
\usepackage{color}
\usepackage{rotating}
\usepackage{overpic}
\usepackage[
   pagebackref=false,
   breaklinks=true,
   letterpaper=true,
   colorlinks,
   bookmarks=false]{hyperref}

%% file: include/nmg/abbrv.tex
\newcommand{\degg}{^\circ{}}              
\def\dash{-\phantom{00}}                  
\def\Hline{\Xhline{2\arrayrulewidth}}     

%% file: include/nmg/authors.tex
\author{
   Tsung-Wei Ke\\
   UC Berkeley / ICSI\\
   {\tt{\small{twke@icsi.berkeley.edu}}}
   \and
   Michael Maire\\
   TTI Chicago\\
   {\tt{\small{mmaire@ttic.edu}}}
   \and
   Stella X. Yu\\
   UC Berkeley / ICSI\\
   {\tt{\small{stellayu@berkeley.edu}}}
}

%% file: include/nmg/sec0_abstract.tex
We propose a multigrid extension of convolutional neural networks (CNNs).
Rather than manipulating representations living on a single spatial grid, our
network layers operate across scale space, on a pyramid of grids.  They
consume multigrid inputs and produce multigrid outputs; convolutional filters
themselves have both within-scale and cross-scale extent.  This aspect is
distinct from simple multiscale designs, which only process the input at
different scales.  Viewed in terms of information flow, a multigrid network
passes messages across a spatial pyramid.  As a consequence, receptive field
size grows exponentially with depth, facilitating rapid integration of context.
Most critically, multigrid structure enables networks to learn internal
attention and dynamic routing mechanisms, and use them to accomplish tasks on
which modern CNNs fail.

Experiments demonstrate wide-ranging performance advantages of multigrid.  On
CIFAR and ImageNet classification tasks, flipping from a single grid to
multigrid within the standard CNN paradigm improves accuracy, while being
compute and parameter efficient.  Multigrid is independent of other
architectural choices; we show synergy in combination with residual
connections.  Multigrid yields dramatic improvement on a synthetic semantic
segmentation dataset.  Most strikingly, relatively shallow multigrid networks
can learn to directly perform spatial transformation tasks, where, in contrast,
current CNNs fail.  Together, our results suggest that continuous evolution of
features on a multigrid pyramid is a more powerful alternative to existing CNN
designs on a flat grid.

%% file: include/nmg/sec1_introduction.tex
\section{Introduction}
\label{sec:introduction}

\begin{figure*}
   \begin{center}
      \includegraphics[width=0.99\linewidth]{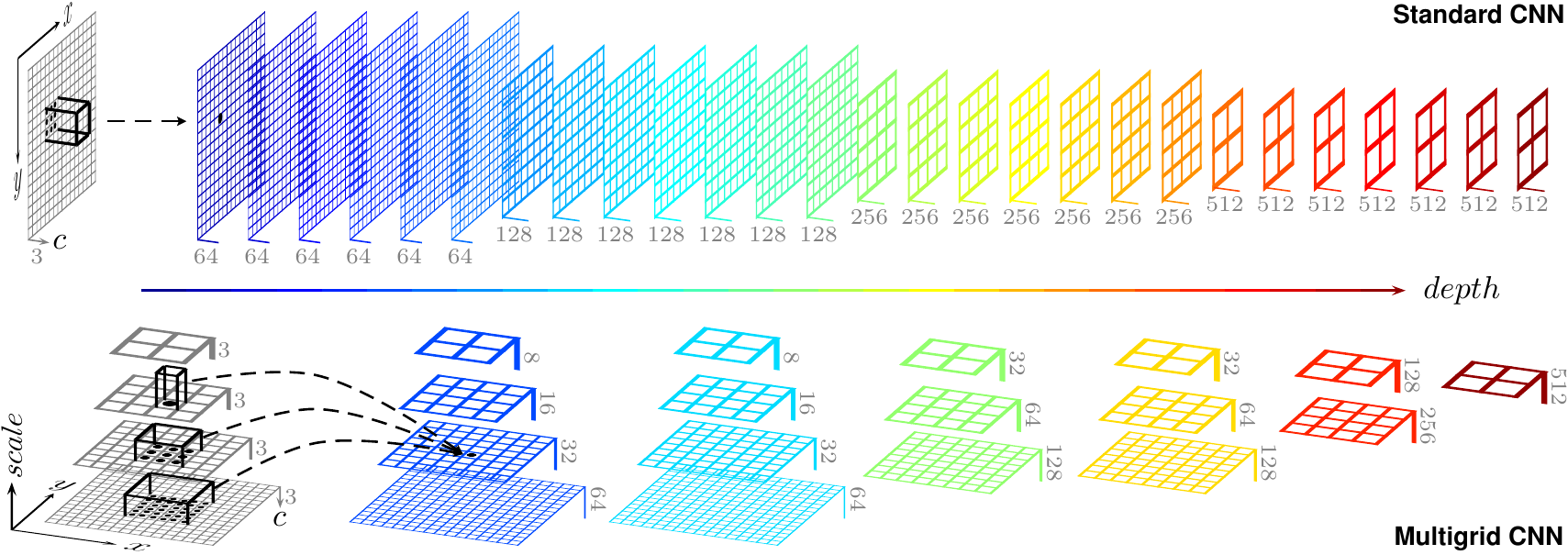}
   \end{center}
   \caption{
      \textbf{Multigrid networks.}
      \emph{\textbf{Top:}}
      Standard CNN architectures conflate scale with abstraction (depth).
      Filters with limited receptive field propagate information slowly across
      the spatial grid, necessitating the use of very deep networks to fully
      integrate contextual cues.
      \emph{\textbf{Bottom:}}
      In multigrid networks, convolutional filters act across scale space
      $(x,y,c,s)$, thereby providing a communication mechanism between coarse
      and fine grids.  This reduces the required depth for mixing distant
      contextual cues to be logarithmic in spatial separation.  Additionally,
      the network is free to disentangle scale and depth: every layer may learn
      several scale-specific filter sets, choosing what to represent on each
      pyramid level.  Traditional pooling and subsampling are now similarly
      multigrid, reducing the size of an entire pyramid.
   }
   \label{fig:multigrid_cnn}
\end{figure*}

Since Fukushima's neocognitron~\cite{neocognitron}, the basic architectural
design of convolutional neural networks has persisted in form similar to that
shown in the top of Figure~\ref{fig:multigrid_cnn}.  Processing begins
on a high resolution input, of which filters examine small local pieces.
Through stacking many layers, in combination with occasional pooling and
subsampling, receptive fields slowly grow with depth, eventually encompassing
the entire input.  Work following this mold includes LeNet~\cite{LeNet}, the
breakthrough AlexNet~\cite{alexNet12}, and the many architectural enhancements
that followed, such as VGG~\cite{vgg16}, GoogLeNet~\cite{szegedy2015inception},
residual networks~\cite{he2015deep, he2016identity, wideresnet} and
like~\cite{fractalnet, densenet}.  Coupled with large datasets and compute
power, this pipeline drives state-of-the-art vision systems.

However, sufficiency of this design does not speak to its optimality.  A
revolution in performance may have blinded the community to investigating
whether or not unfolding computation in this standard manner is the best
choice.  In fact, there are shortcomings to the typical CNN pipeline:
\begin{itemize}
   \item{
      It conflates abstraction and scale.  Early layers cannot see coarser
      scales, while later layers only see them.  For tasks requiring
      fine-scale output, such as semantic segmentation, this necessitates
      specialized designs for reintegrating spatial information~\cite{
      long_shelhamer_fcn, HAGM:CVPR:2015, DeepLab:2016, SegNet}.
   }
   \item{
      The fine-to-coarse processing within a standard CNN is in opposition to a
      near universal principle for efficient algorithm design: coarse-to-fine
      processing.  The first layer in a standard CNN consists of many filters
      independently looking at tiny, almost meaningless regions of the image.
      Would it not be more reasonable for the system to observe some
      coarse-scale context before deciding how to probe the details?
   }
   \item{
      Communication is inefficient.  A neuron's receptive field is determined
      by the units in the input layer that could propagate a signal to it.
      Standard CNNs implement a slow propagation scheme, diffusing information
      across a single grid at rate proportional to convolutional filter size.
      This may be a reason extremely deep networks~\cite{
      szegedy2015inception, he2015deep, fractalnet} appear necessary; many
      layers are needed to counteract inefficient signal propagation.
}
\end{itemize}

These points can be summarized as inherent deficiencies in representation,
computation, and communication.  Our multigrid architecture
(Figure~\ref{fig:multigrid_cnn}, bottom) endows CNNs with additional structural
capacity in order to dissolve these deficiencies.  It is explicitly multiscale,
pushing choices about scale-space representation into the training process.

Computation occurs in parallel at all scales; every layer process both coarse
and fine representations.  Section~\ref{sec:progressive} also explores
coarse-to-fine variants that transition from processing on a coarse pyramid to
processing on a full pyramid as the network deepens.  Pyramids provide not only
an efficient computational model, but a unified one.  Viewing the network as
evolving a representation living on the pyramid, we can combine previous
task-specific architectures.  For classification, attach an output to the
coarsest pyramid level; for segmentation, attach an output to the finest.

Multigrid structure facilitates cross-scale information exchange, thereby
destroying the long-established notion of receptive field.  Most neurons have
receptive field equivalent to the entire input; field size grows exponentially
with depth, or, in progressive multigrid networks, begins with the full
(coarse) input.  Quick communication pathways exist throughout the network,
and enable new capabilities.

We specifically demonstrate that multigrid CNNs, trained in a pure end-to-end
fashion, can learn to attend and route information.  Their emergent behavior
may dynamically emulate the routing circuits articulated by
Olshausen~\etal~\cite{Olshausen93}.  We construct a synthetic task that
standard CNNs completely fail to learn, but multigrid CNNs accomplish with
ease.  Here, attentional capacity is key.

As Section~\ref{sec:related} reviews, recent CNN architectural innovations
ignore scale-space routing capacity, focusing instead on aspects like depth.
Multigrid, as Section~\ref{sec:multigrid} details, complements such work.
Section~\ref{sec:experiments} measures performance improvements due to
multigrid on classification tasks (CIFAR and ImageNet) and synthetic semantic
segmentation tasks.  Multigrid boosts both baseline and residual CNNs.  On a
synthetic spatial transformation task, multigrid is more than a boost; it
is required, as residual networks alone do not possess attentional capacity.
Section~\ref{sec:conclusion} discusses implications.

\begin{figure*}
   \begin{center}
      \includegraphics[angle=90,width=0.99\linewidth]{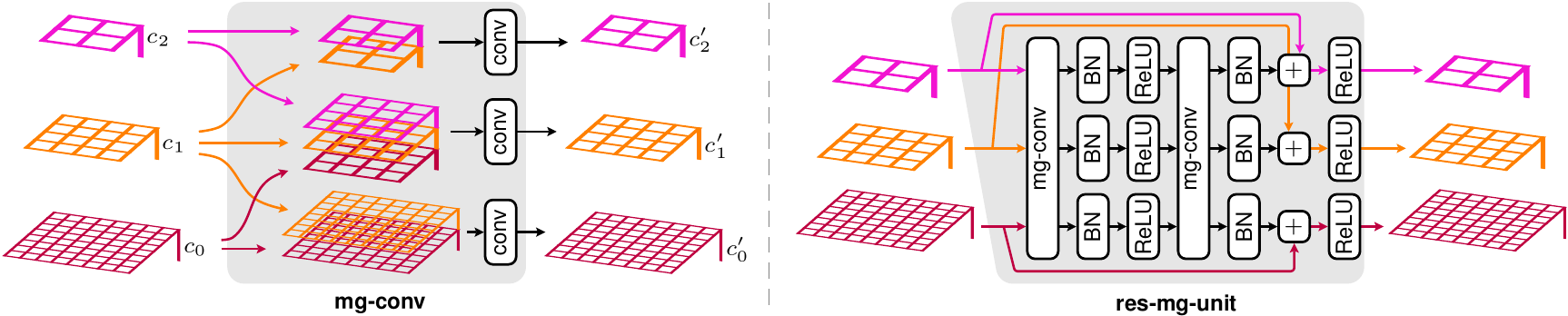}
   \end{center}
   \vspace{-0.01\linewidth}
   \caption{
      \textbf{Multigrid layers.}
      \emph{\textbf{Left:}}
      We implement a multigrid convolutional layer
      ({\footnotesize{\textsf{mg-conv}}}) using readily available components.
      For each grid in the input pyramid, we rescale its neighboring grids to
      the same spatial resolution and concatenate their feature channels.
      Convolution over the resulting single grid approximates scale-space
      convolution on the original pyramid.  Downscaling is max-pooling, while
      upscaling is nearest-neighbor interpolation.
      \emph{\textbf{Right:}}
      A building block ({\footnotesize{\textsf{res-mg-unit}}}) for residual
      connectivity~\cite{he2015deep} in multigrid networks.
   }
   \label{fig:multigrid_layers}
\end{figure*}

%% file: include/nmg/sec2_related.tex
\section{Related Work}
\label{sec:related}

In wake of AlexNet~\cite{alexNet12}, exploration of CNNs across computer vision
has distilled some rules of thumb for their design.  Small (\eg~$3 \times 3$)
spatial filters, in many successive layers, make for efficient parameter
allocation~\cite{vgg16, szegedy2015inception, he2015deep}.  Feature channels
should increase with spatial resolution reduction~\cite{alexNet12,vgg16}
(\eg~doubling as in Figure~\ref{fig:multigrid_cnn}).  Deeper networks are
better, so long as a means of overcoming vanishing gradients is engineered into
the training process~\cite{szegedy2015inception,huang2016stochasticdepth} or
the network itself~\cite{srivastava2015highway,he2015deep,fractalnet,densenet}.
Width matters~\cite{wideresnet}.

The desire to adapt image classification CNNs to more complex output tasks,
such as semantic segmentation, has catalyzed development of ad-hoc
architectural additions for restoring spatial resolution.  These include skip
connections and upsampling~\cite{long_shelhamer_fcn,DeepLab:2016},
hypercolumns~\cite{HAGM:CVPR:2015}, and, autoencoder-like, hourglass or
U-shaped networks that reduce and re-expand spatial grids~\cite{SegNet,CLVM,
hourglass,UNet}.

This latter group of methods reflects the classic intuition of connecting
bottom-up and top-down signals.  Our work differs from these and earlier
models~\cite{nap} by virtual of decoupling pyramid level from feature
abstraction.  Representations at all scales evolve over the depth of our
network.  Such dynamics also separates us from past multiscale CNN
work~\cite{Farabet,EPF:NIPS:2014,XT:ICCV:2015}, which does not consider
embedded and continual cross-scale communication.

While prior CNN designs have varied filter size (\eg~\cite{alexNet12,
szegedy2015inception}), our choice to instead vary grid resolution is
crucial.  Applying multiple filter sets, of different spatial size, to a
single grid could emulate a multigrid computation.  However, we use exponential
(power of $2$) grid size scaling.  Emulating this via larger filters quickly
becomes cost prohibitive in terms of both parameters and computation; it is
impractical to implement in the style of Inception modules~\cite{
szegedy2015inception}.  Dilated~\cite{msDilated} and atrous convolution~\cite{
DeepLab:2016}, while related, do not fully capture the pooling and
interpolation aspects (detailed in Section~\ref{sec:multigrid}) of our
multigrid operator.

Recent efforts to improve the computational efficiency of CNNs, though too
numerous to list in full, often attack via parameter~\cite{DeepFried,
CompHash, SqueezeNet} or precision~\cite{XNORnet, BinaryConnect} reduction.
Concurrent work extends such exploration to include cascades~\cite{msDense,
MP:arXiv:2017}, in which explicit routing decisions allow for partial
evaluation of a network.  Unlike this work, our cross-scale connections are
bi-directional and we explore routing in a different sense: internal transport
of information (\eg~attention~\cite{Olshausen93}) for solving a particular
task.  We focus on a coarse-to-fine aspect of efficiency, borrowing from the
use of multigrid concepts in image segmentation~\cite{SGSBB:Nature:2006,
MY:ICCV:2013}.

%% file: include/nmg/sec3_multigrid.tex
\section{Multigrid Architectures}
\label{sec:multigrid}

\begin{figure*}
   \begin{center}
      \includegraphics[width=0.99\linewidth]{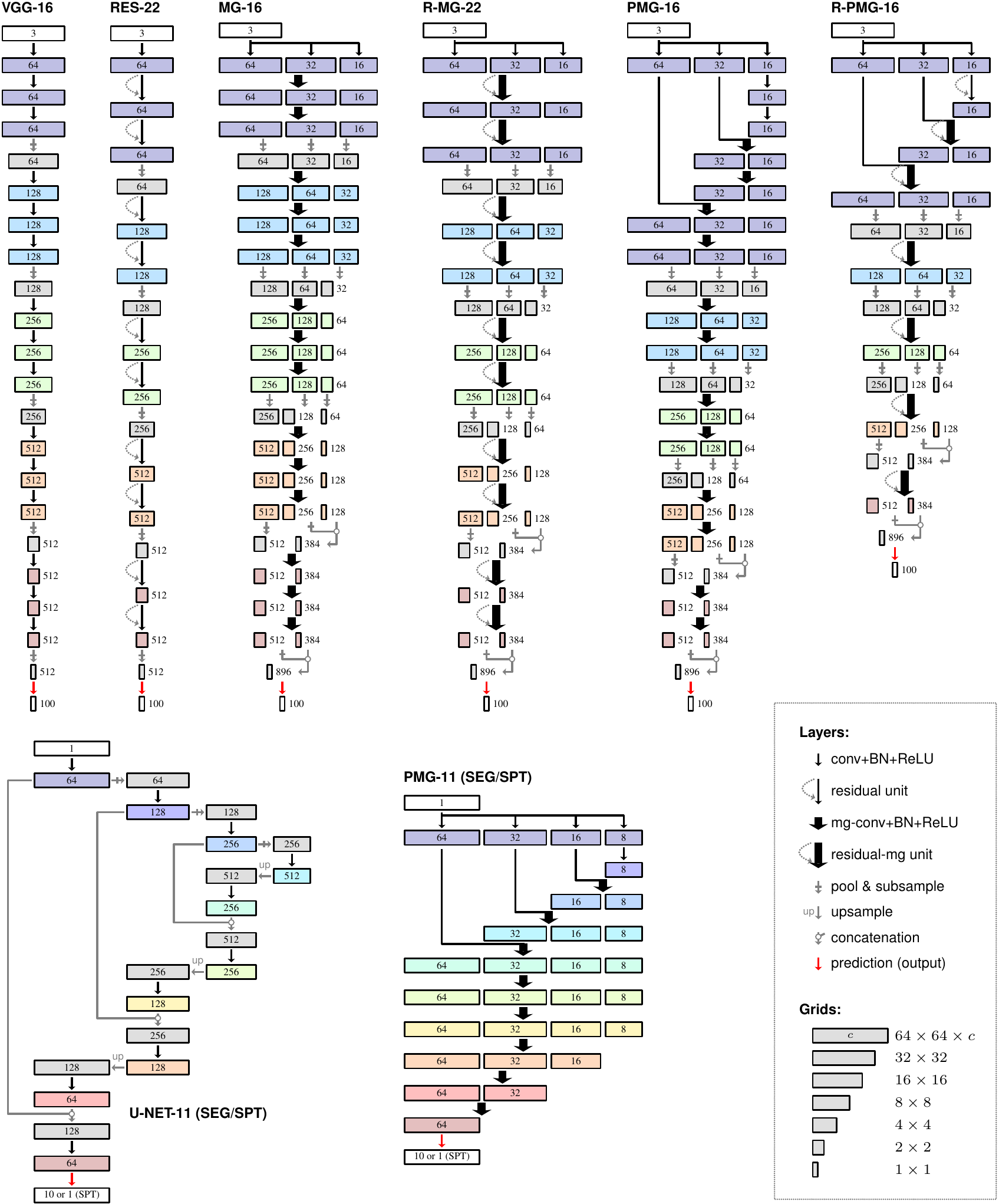}
   \end{center}
   \vspace{-0.005\linewidth}
   \caption{
      \textbf{Network diagrams.}
      \emph{\textbf{Top:}}
         Baseline (VGG) and multigrid (MG) CNNs for CIFAR-100 classification,
         as well as residual (RES, R-) and progressive (PMG) versions.  We vary
         depth by duplicating layers within each colored block.
      \emph{\textbf{Bottom:}}
         Semantic segmentation (SEG) and spatial transformation (SPT)
         architectures: U-NET~\cite{UNet}, and our progressive multigrid
         alternative (PMG).  We also consider U-MG (not shown), a baseline
         that replaces U-NET grids with pyramids (as in VGG $\rightarrow$ MG),
         yet still shrinks and re-expands pyramids.
   }
   \label{fig:nets}
\end{figure*}

Figure~\ref{fig:multigrid_cnn} conveys our intention to wire cross-scale
connections into network structure at the lowest level.  We can think of a
multigrid CNN as a standard CNN in which every grid is transformed into a
pyramid.  Every convolutional filter extends spatially within grids $(x,y)$,
across grids of multiple scales $(s)$ within a pyramid, and over corresponding
feature channels $(c)$.  A pyramidal slice, across scale-space, of the
preceding layer contributes to the response at a particular corresponding
neuron in the next.

This scheme enables any neuron to transmit a signal up the pyramid, from fine
to coarse grids, and back down again.  Even with signals jumping just one
pyramid level per consecutive layer, the network can exploit this structure to
quickly route information between two spatial locations on the finest grid.
Communication time is logarithmic in spatial separation, rather than linear as
in standard CNNs.

This difference in information routing capability is particularly dramatic in
light of the recent trend towards stacking many layers of small $3 \times 3$
convolutional filters~\cite{vgg16,he2015deep}.  In standard CNNs, this
virtually guarantees that either very deep networks, or manually-added pooling
and unpooling stages~\cite{SegNet,CLVM,hourglass,UNet}, will be needed to
propagate information across the pixel grid.  Multigrid allows for faster
propagation with minimal additional design complexity.  Moreover, unlike fixed
pooling/unpooling stages, multigrid allows the network to learn data-dependent
routing strategies.

Instead of directly implementing multigrid convolution as depicted in
Figure~\ref{fig:multigrid_cnn}, we implement a close alternative that can be
easily built out of standard components in existing deep learning frameworks.
Figure~\ref{fig:multigrid_layers} illustrates the multigrid convolutional
layer ({\small{\textsf{mg-conv}}}) we use as a drop-in replacement for standard
convolutional ({\small{\textsf{conv}}}) layers when converting a network from
grids to pyramids.

With multigrid convolution, we may choose to learn independent filter sets for
each scale within a layer, or alternatively, tie parameters between scales and
learn shared filters.  We learn independent sets, assuming that doing so
affords the network the chance to squeeze maximum performance out of
scale-specific representations.  Channel counts between grids need not match
in any particular manner.  Each of $c_0,c_1,c_2,c_0',c_1',c_2'$ in
Figure~\ref{fig:multigrid_layers} is independent.  For comparisons, we either
set channel count on the finest grid to match baseline models, or calibrate
channel counts so that total parameters are similar to the baselines.
Section~\ref{sec:experiments} reveals quite good performance is achievable by
halving channel count with each coarser grid (\eg~$c_1 = \frac{1}{2}c_0$,
$c_2 = \frac{1}{4}c_0$).  Thus, multigrid adds minimal computational overhead;
the coarse grids are cheap.

Given recent widespread use of residual networks~\cite{he2015deep}, we consider
a multigrid extension of them.  The right side of
Figure~\ref{fig:multigrid_layers} shows a multigrid analogue of the original
residual unit~\cite{he2015deep}.  Convolution acts jointly across the multigrid
pyramid, while batch normalization (BN)~\cite{batchnorm} and ReLU apply
separately to each grid.  Extensions are also possible with alternative
residual units~\cite{he2016identity}, but we leave such exploration to future
work.  Our focus is on comparison of baseline, residual, multigrid, and
residual-multigrid networks.

\subsection{Pooling}
\label{sec:pooling}

Within an {\small{\textsf{mg-conv}}} layer, max-pooling and subsampling acts
as a lateral communication mechanism from fine to coarse grids.  Similarly,
upsampling facilitates lateral communication from coarse to fine grids.  Rather
than locating these operations at a few fixed stages in the pipeline, we are
actually inserting them everywhere.  However, their action is now lateral,
combining different-scale grids in the same layer, rather than between grids of
different scales in two layers consecutive in depth.

While Figure~\ref{fig:multigrid_cnn} shows additional max-pooling and
subsampling stages acting depth-wise on entire pyramids, we also consider
pipelines that do not shrink pyramids.  Rather, they simply attach an output to
a particular grid (and possibly prune a few grids outside of its communication
range).  Success of this strategy motivates rethinking the role of pooling in
CNNs.  Instead of explicit summarization employed at select stages, multigrid
yields a view of pooling as implicit communication that pervades the network.

\subsection{Progressive Multigrid}
\label{sec:progressive}

The computation pattern underlying a multigrid CNN's forward pass has analogy
with the multiscale multigrid scheme of Maire and Yu~\cite{MY:ICCV:2013}.  One
can view their eigensolver as a linear diffusion process on a multiscale
pyramid.  We view a multigrid CNN as a nonlinear process, on a similar pyramid,
with the same communication structure.

Extending the analogy, we port their progressive multigrid computation scheme
to our setting.  Rather than starting directly on the full pyramid, a
progressive multigrid CNN can spend several layers processing only a coarse
grid.  Following this are additional layers processing a small pyramid
(\eg~coarse and medium grids together), before the remainder of the network
commits to work with the full pyramid.

In all multigrid and progressive multigrid experiments, we set the fine-scale
input grid to be the original image and simply feed in downsampled (lower
resolution) versions to independent initial convolution layers.  The outputs
of these initial layers then form a multigrid pyramid which is processed
coherently by the rest of the network.

\subsection{Model Zoo}
\label{sec:zoo}

Figure~\ref{fig:nets} diagrams the variety of network architectures we
evaluate.  For classification, we take a network with minor differences from
that of Simonyan and Zisserman~\cite{vgg16} as our baseline.  In abuse of
notation, we reuse their VGG name.  Our 16-layer version, VGG-16, consists of
5 sections of 3 convolutional layers each, with pooling and subsampling
between.  A final softmax layer produces class predictions (on CIFAR-100).
Convolutional layers use $3 \times 3$ filters.  We instantiate variants
with different depth by changing the number layers per section (2 for
VGG-11, 4 for VGG-21, \etc).  Residual baselines (RES) follow the same layout,
but use residual units within each section.  Recall that each residual unit
contains two convolutional layers.

Multigrid (MG) and residual multigrid (R-MG) networks, starting respectively
from VGG or RES, simply flip from grids to pyramids and convolution to
multigrid convolution.  Progressive variants (PMG, R-PMG) expand the first
section in order to gradually work up to computation on the complete pyramid.
Even as their depth increases, a significant portion of layers in progressive
networks avoid processing the full pyramid.  As diagrammed, multigrid networks
match their fine-grid feature count with baselines, and hence add some capacity
and parameters.  For classification, we also consider smaller (denoted -sm)
multigrid variants with fewer channels per grid in order to be closer in
parameter count with baseline VGG and RES networks.

For the semantic segmentation and spatial transformation tasks detailed in the
next section, we use networks that produce per-pixel output.  As a baseline,
we employ the U-NET design of Ronneberger~\etal~\cite{UNet}.  Again,
convolutional filters are $3 \times 3$, except for the layers immediately
following upsampling; here we use $2 \times 2$ filters, following~\cite{UNet}.
We examine progressive multigrid alternatives (PMG, R-PMG) that continue
to operate on the multiscale pyramid.  Unlike the classification setting, we
do not reduce pyramid resolution.  These networks drop some coarse grids
towards the end of their pipelines for the sole reason that such grids do not
communicate with the output.

%% file: include/nmg/sec4_experiments.tex
\begin{table}[t]
   \definecolor{darkgreen}{rgb}{0,0.5,0}
   \begin{center}
      \begin{footnotesize}
         \input{include/nmg/tab_cifar100.tex}
      \end{footnotesize}
      \vspace{0.01\linewidth}
      \caption{
         \textbf{CIFAR-100 classification performance.}
      }
      \label{tab:cifar100_classify}
   \end{center}
   \vspace{-0.075\linewidth}
\end{table}

\section{Experiments}
\label{sec:experiments}

Our experiments focus on systematic exploration and evaluation of the
architectural design space, with the goal of quantifying the relative
benefits of multigrid and its synergistic combination with residual
connections.

\subsection{Classification: CIFAR-100~\cite{CIFAR}}
\label{sec:classify}

We evaluate the array of network architectures listed in
Table~\ref{tab:cifar100_classify} on the task of CIFAR-100 image
classification.

\vspace{-0.03\linewidth}
~\\\noindent
{\bf Data}.
We whiten and then enlarge images to $36\times36$.  We use random $32\times32$
patches from the training set (with random horizontal flipping) and the center
$32\times32$ patch from test examples.  There are 50K training and 10K test
images.

\vspace{-0.03\linewidth}
~\\\noindent
{\bf Training.}
We use SGD with batch size 128, weight decay rate 0.0005, and 300 iterations
per epoch for 200 epochs.  For VGG, MG, and PMG, learning rate exponentially
decays from 0.1 to 0.0001; whereas for RES, R-MG, and R-PMG, it decays at rate
0.2 every 60 epochs.

\vspace{-0.03\linewidth}
~\\\noindent
{\bf Results.}
From Table~\ref{tab:cifar100_classify}, we see that adding multigrid
capacity consistently improves over both basic (compare green entries),
as well as residual CNNs (blue entries).  Progressive multigrid nets of similar
depth achieve better or comparable accuracy at reduced expense, as quantified
in terms of both parameters and floating point operations (red entries).

\subsection{Semantic Segmentation}
\label{sec:seg}

We generate synthetic data (10K training, 1K test images) for semantic
segmentation by randomly scaling, rotating, translating, and pasting
MNIST~\cite{LeNet} digits onto a $64 \times 64$ canvas, limiting digit overlap
to $30\%$.  The task is to decompose the result into per-pixel digit class
labels.

In addition to networks already mentioned, we consider a single grid (SG)
baseline that mirrors PMG but removes all grids but the finest.  We also
consider U-MG, which changes resolution like U-NET, but works on pyramids.

\begin{table}
   \begin{center}
      \begin{footnotesize}
         \input{include/nmg/tab_mnist_seg.tex}
      \end{footnotesize}
      \vspace{0.01\linewidth}
      \caption{
         \textbf{MNIST semantic segmentation performance.}
      }
      \label{tab:mnist_seg}
   \end{center}
   \vspace{-0.05\linewidth}
\end{table}

\begin{table}
   \begin{center}
      \begin{footnotesize}
         \input{include/nmg/tab_mnist_spt.tex}
      \end{footnotesize}
      \vspace{0.01\linewidth}
      \caption{
         \textbf{MNIST spatial transformation performance.}
      }
      \label{tab:mnist_spt}
   \end{center}
   \vspace{-0.05\linewidth}
\end{table}

\begin{table}[t]
   \begin{center}
      \begin{footnotesize}
         \setlength\tabcolsep{3pt}
         \input{include/nmg/tab_imagenet.tex}
      \end{footnotesize}
      \vspace{-0.005\linewidth}
      \caption{
         \textbf{ImageNet classification error (\%) on validation set.}
      }
      \label{tab:imagenet}
   \end{center}
   \vspace{-0.08\linewidth}
\end{table}

\begin{figure*}
   \begin{center}
      \ifarxiv
         \input{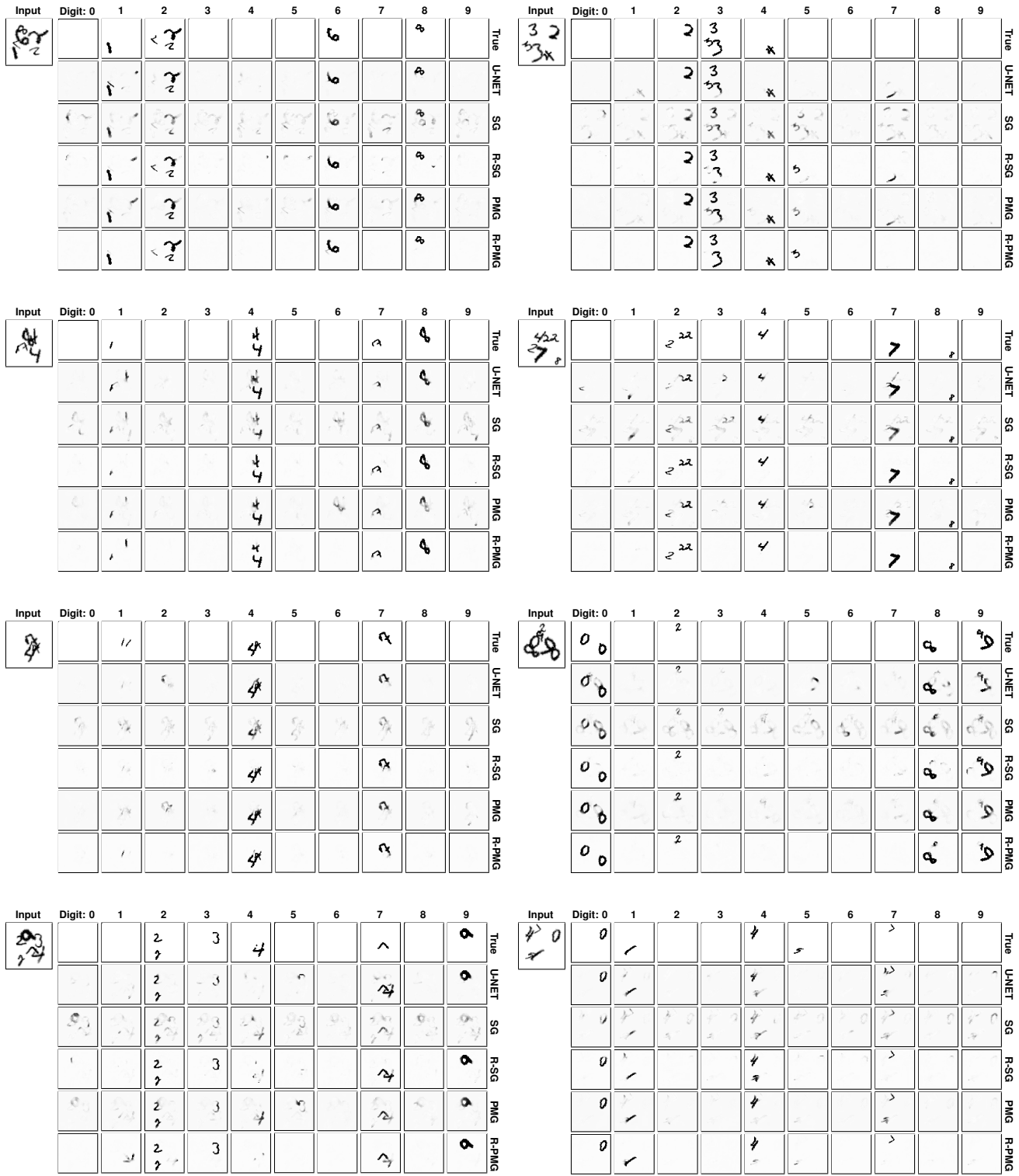}
      \fi
   \end{center}
   \ifarxiv\vspace{0.02\linewidth}\fi
   \caption{
      \textbf{Semantic segmentation on synthetic MNIST.}
      We compare U-NET, residual (R-), and progressive multigrid (PMG)
      networks on a synthetic task of disentangling randomly distorted and
      overlapped MNIST digits.  Figure~\ref{fig:nets} specifies the $11$-layer
      U-NET and PMG variants used here.  An $11$-layer single-grid (SG)
      baseline operates on only the finest scale.  Analogous residual versions
      (R-SG, R-PMG) are $20$ layers deep.  PMG thus performs similarly
      to a deeper single-grid residual network (R-SG), while R-PMG bests all.
   }
   \label{fig:seg}
\end{figure*}

\vspace{-0.03\linewidth}
~\\\noindent
{\bf Training.}
We use batch size 64, weight decay rate 0.0005, and 150 iterations per epoch
for 200 epochs.  We set learning rate schedules for non-residual and
residual networks in the same manner as done for CIFAR-100.

\vspace{-0.03\linewidth}
~\\\noindent
{\bf Results.}
Table~\ref{tab:mnist_seg} and Figure~\ref{fig:seg} show dramatic performance
advantages of the progressive multigrid network.  We report scores of mean
intersection over union (IoU) with the true output regions, as well as mean
per-pixel multiclass labeling accuracy.  These results are a strong indicator
that continuous operation on a multigrid pyramid should replace networks that
pass through a low resolution bottleneck.

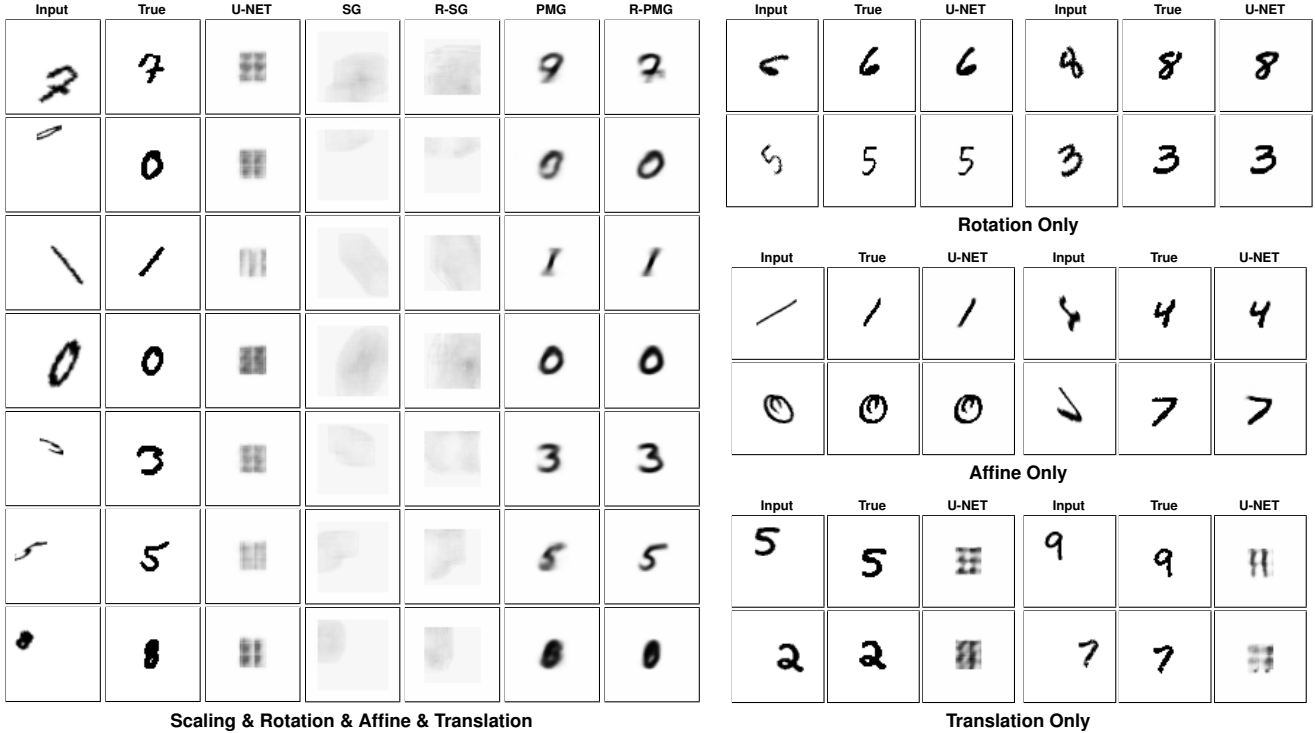
\begin{figure*}
   \vspace{0.005\linewidth}
   \begin{center}
      \input{include/nmg/fig_spt_content.tex}
   \end{center}
   \vspace{-0.005\linewidth}
   \caption{
      \textbf{Spatial transformations on synthetic MNIST.}
      \emph{\textbf{Left:}}
         Only multigrid CNNs (PMG, R-PMG) are able to learn to take a
         distorted and translated digit as input and produce the centered
         original.
      \emph{\textbf{Right:}}
         U-NET can learn to undo rotation and affine transformations, but not
         translation.  U-NET's failure stems from inability to properly attend
         and route information from a sub-region of the input.
   }
   \label{fig:spt}
\end{figure*}

\begin{figure*}
   \vspace{0.005\linewidth}
   \begin{center}
      \input{include/nmg/fig_sal_content.tex}
   \end{center}
   \vspace{-0.005\linewidth}
   \caption{
      \textbf{Attention maps for MNIST spatial transformers.}
      To produce attention maps, we sweep an occluder ($8 \times 8$ square of
      uniform random noise) over the $64\times 64$ input image and measure
      sensitivity of the output at the red and green locations.  U-NET, which
      fails to learn translation, attends to approximately the same regions
      regardless of input.  PMG and R-PMG instead exhibit attention that
      translates with the input.
      \ifarxiv
         See Figure~\ref{fig:pipeline} for a detailed visualization of our
         attention map generation process.
      \fi
   }
   \label{fig:sal}
\end{figure*}
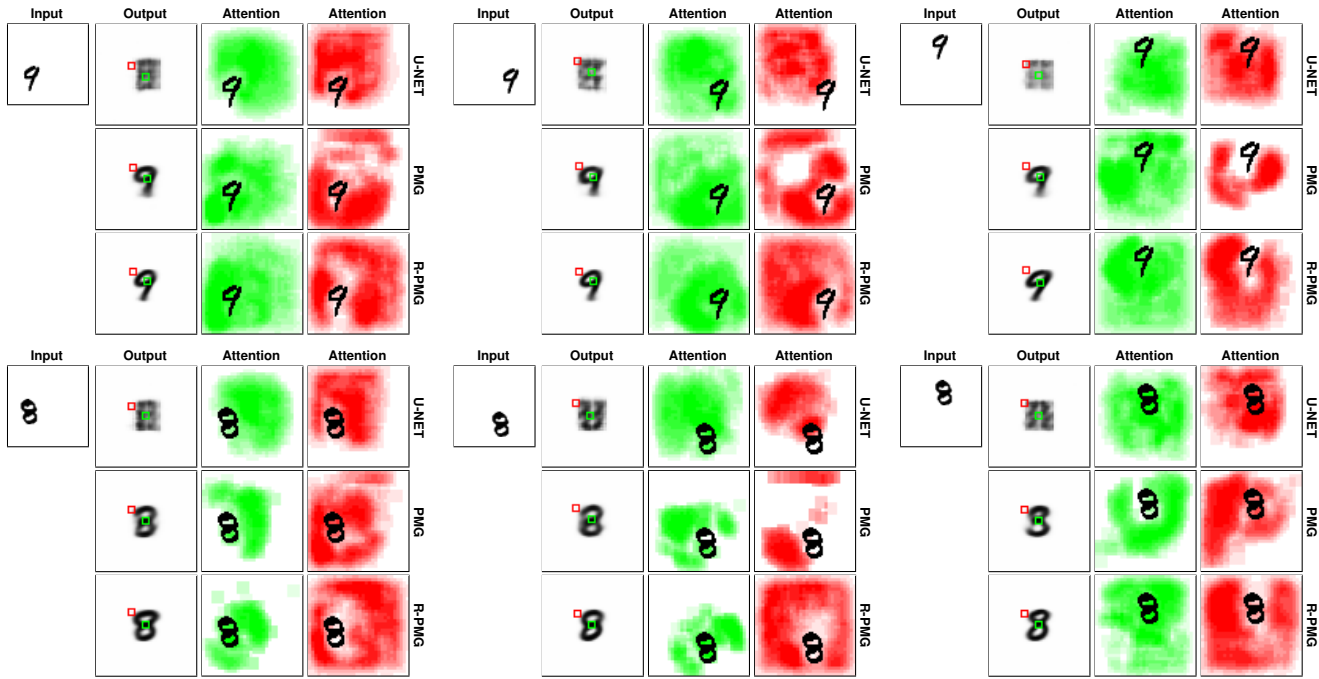

\subsection{Spatial Transformers}
\label{sec:spt}

Jaderberg~\etal~\cite{SPT} engineer a system for inverting spatial distortions
by combining a neural network for estimating the parameters of the
transformation with a dedicated unit for applying the inverse.  Such a split
of the problem into two components appears necessary when using standard CNNs;
they cannot learn the task end-to-end.  But, progressive multigrid networks
are able to learn such tasks.

We generate a synthetic spatial transformation dataset (60K training, 10K
test images) in the same manner as the one for semantic segmentation, but with
only one digit per image and an additional affine transformation with a
uniformly random sheering angle in $(-60\degg{},60\degg{})$.  Training
is the same as segmentation, except 800 iterations per epoch.

\vspace{-0.03\linewidth}
~\\\noindent
{\bf Results.}
Table~\ref{tab:mnist_spt} and Figure~\ref{fig:spt} show that all networks
except PMG and R-PMG fail to learn the task.  Figure~\ref{fig:spt} (right
side) reveals the reason: U-NET (and others not shown) cannot learn
translation.  This makes sense for single grid methods, as propagating
information across the fine grid would require them to be deeper than tested.
U-NET's failure seems to stem from confusion due to pooling/upsampling.
It appears to paste subregions of the target digit into output, but not in the
correct arrangement.  Figure~\ref{fig:sal} reveals that, unlike U-NET, PMG and
R-PMG exhibit attentional behavior.

\ifarxiv
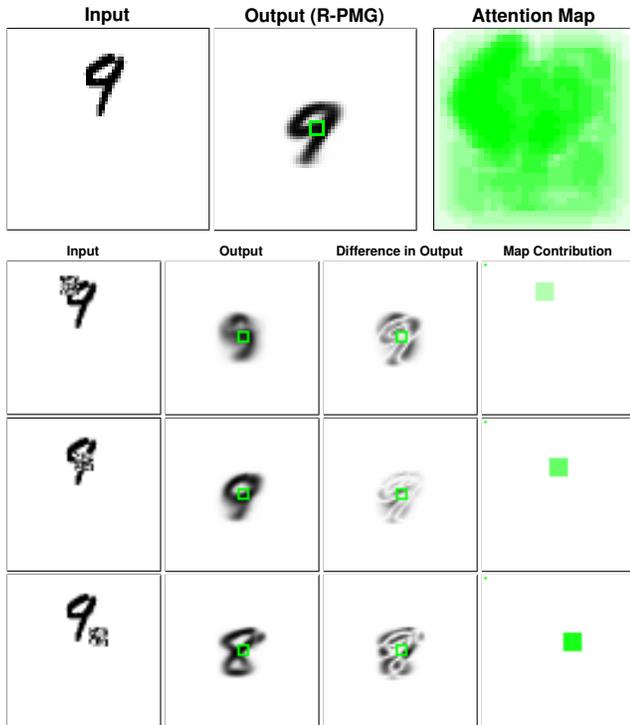
\begin{figure}
   \vspace{0.005\linewidth}
   \begin{center}
      \input{include/nmg/fig_sal_pipeline.tex}
   \end{center}
   \vspace{-0.005\linewidth}
   \caption{
      \textbf{Attention map generation.}
      For a particular probe location, \eg~open green square in the center, we
      compute an attention map as follows.  We slide an occluder, consisting of
      an $8 \times 8$ square of uniform random noise, across the input, and
      compute a corresponding output image for each position of the occluder.
      The maximum absolute difference between this output and the original
      output, in a $3 \times 3$ region around the probe, defines the magnitude
      of attention to the occluder (equivalently, the influence of the
      occluder on output at the probe).  Combining contributions from all
      occluders yields an entire attention map.
   }
   \label{fig:pipeline}
\end{figure}
\fi

\subsection{Classification: ImageNet~\cite{deng2009imagenet}}
\label{sec:imagenet}

We select two designs to train from scratch on ImageNet:
\begin{itemize}
   \item{
      \vspace{-0.01\linewidth}
      R-MG-34: A multigrid analogue of ResNet-34~\cite{he2015deep}, with
      finest grid matching ResNet, and two additional grids at half and quarter
      spatial resolution (and channel count).  Pooling and resolution
      reduction mirror ResNet-34 and the R-MG diagram in Figure~\ref{fig:nets}.
   }
   \item{
      \vspace{-0.02\linewidth}
      R-PMG-30-SEG: Similar to our semantic segmentation network, but adapted
      to ImageNet, it maps $224 \times 224$ input onto 4 grids: $56 \times 56$,
      $28 \times 28$, $14 \times 14$, $7 \times 7$, that are maintained
      throughout the network without pooling stages.  We increase feature
      count with depth and attach a classifier to the final coarsest grid.
   }
\end{itemize}
Training uses SGD with batch size 256, weight decay 0.0001, 10K iterations
per epoch for 100 epochs, starting learning rate 0.1 with decay rate 0.1 every
30 epochs.  Table~\ref{tab:imagenet} compares our performance to that of
ResNets and wide residual networks (WRN)~\cite{wideresnet}.  We observe:
\begin{itemize}
   \item{
      \vspace{-0.01\linewidth}
      Multigrid substantially improves performance.  R-MG-34 even outperforms
      the deeper ResNet-50.
   }
   \item{
      \vspace{-0.02\linewidth}
      Multigrid is a more efficient use of parameters than simply increasing
      feature channel count on a single grid.  Compare R-MG-34 with the
      double-wide residual network WRN-34: we match its performance with fewer
      parameters and require drastically fewer FLOPs.
   }
   \item{
      \vspace{-0.02\linewidth}
      R-PMG-30-SEG outperforms ResNet-34.  The same design we used for
      segmentation is great for classification, lending support to the idea
      of a single unifying architecture which evolves a multigrid
      representation.
   }
\end{itemize}

%% file: include/nmg/tab_cifar100.tex
\begin{tabular}{@{}l|r|r|r@{}}
Method $\qquad\qquad$& Params ($\times 10^6$) & FLOPs ($\times 10^6$) & Error (\%) \\
\Hline
VGG-6 & 3.96 & 67.88 & 32.15\\
\textcolor{darkgreen}{VGG-11} & \textcolor{darkgreen}{9.46} & \textcolor{darkgreen}{228.31} & \textcolor{darkgreen}{30.26}\\
\textcolor{darkgreen}{VGG-16} & \textcolor{darkgreen}{14.96} & \textcolor{darkgreen}{388.75} & \textcolor{darkgreen}{30.47}\\
VGG-21 & 20.46 & 549.18 & 31.89\\
\hline
MG-sm-6 & 3.28 & 46.10 & 33.83\\
MG-sm-11 & 8.02 & 154.60 & 31.55\\
\textcolor{red}{MG-sm-16} & \textcolor{red}{12.75} & \textcolor{red}{263.10} & \textcolor{red}{32.59}\\
MG-sm-21 & 17.49 & 371.60 & 34.21\\
\hline
MG-6 & 8.34 & 116.63 & 32.08\\
\textcolor{darkgreen}{MG-11} & \textcolor{darkgreen}{20.46} & \textcolor{darkgreen}{391.88} & \textcolor{darkgreen}{\textbf{28.39}}\\
\textcolor{darkgreen}{MG-16} & \textcolor{darkgreen}{32.58} & \textcolor{darkgreen}{667.13} & \textcolor{darkgreen}{\textbf{29.91}}\\
MG-21 & 44.69 & 942.38 & 30.03\\
\hline
PMG-sm-9 & 3.33 & 73.23 & 32.38\\
\textcolor{red}{PMG-sm-16} & \textcolor{red}{\textbf{8.07}} & \textcolor{red}{\textbf{183.34}} & \textcolor{red}{\textbf{30.47}}\\
PMG-sm-30 & 12.82 & 293.46 & 31.06\\
\hline
PMG-9 & 8.46 & 186.21 & 30.61\\
PMG-16 & 20.60 & 468.09 & 28.11\\
PMG-30 & 32.74 & 749.98 & 29.89\\
\Hline
RES-12 & 9.50 & 266.02 & 28.64\\
\textcolor{blue}{RES-22} & \textcolor{blue}{20.49} & \textcolor{blue}{586.93} & \textcolor{blue}{28.05}\\
\textcolor{blue}{RES-32} & \textcolor{blue}{31.49} & \textcolor{blue}{907.79} & \textcolor{blue}{27.12}\\
RES-42 & 42.48 & 1228.65 & 27.60\\
\hline
R-MG-sm-12 & 8.06 & 180.12 & 30.24\\
R-MG-sm-22 & 17.52 & 397.11 & 28.65\\
R-MG-sm-32 & 26.99 & 614.11 & 29.27\\
\textcolor{red}{R-MG-sm-42} & \textcolor{red}{36.46} & \textcolor{red}{831.11} & \textcolor{red}{\textbf{26.85}}\\
\hline
R-MG-12 & 20.56 & 457.20 & 27.84\\
\textcolor{blue}{R-MG-22} & \textcolor{blue}{44.79} & \textcolor{blue}{1007.70} & \textcolor{blue}{\textbf{26.79}}\\
\textcolor{blue}{R-MG-32} & \textcolor{blue}{69.02} & \textcolor{blue}{1558.20} & \textcolor{blue}{\textbf{25.29}}\\
R-MG-42 & 93.26 & 2108.71 & 26.32\\
\hline
R-PMG-sm-16 & 8.07 & 183.34 & 29.41\\
R-PMG-sm-30 & 17.56 & 403.57 & 27.68\\
\textcolor{red}{R-PMG-sm-44} & \textcolor{red}{\textbf{27.05}} & \textcolor{red}{\textbf{623.79}} & \textcolor{red}{\textbf{27.00}}\\
\hline
R-PMG-16 & 20.60 & 468.09 & 27.35\\
R-PMG-30 & 44.88 & 1031.87 & 26.44\\
R-PMG-44 & 69.16 & 1595.64 & 26.68\\
\hline
\end{tabular}

%% file: include/nmg/tab_mnist_seg.tex
\begin{tabular}{@{}l|r|r|r@{}}
Method $\qquad$& Params ($\times 10^6$) & mean IoU (\%) & mean Error (\%)\\
\Hline
U-NET-11 & 3.79 & 58.46 & 22.14\\
U-MG-11  & 5.90 & 58.02 & 21.45\\
SG-11    & 0.23 & 32.21 & 22.33\\
R-SG-20  & 0.45 & 69.86 & 14.43\\
PMG-11   & 0.61 & 61.88 & 14.75\\
R-PMG-20 & 1.20 & 81.91 &  8.89\\
\hline
\end{tabular}

%% file: include/nmg/tab_mnist_spt.tex
\begin{tabular}{@{}l|r|r|r@{}}
Method $\qquad$& Params ($\times 10^6$) & mean IoU (\%) & mean Error (\%)\\
\Hline
U-NET-11 & 3.79 & 12.20          &  37.55\\
SG-11    & 0.23 & 0              & 100.00\\
R-SG-20  & 0.45 & 0              &  82.98\\
PMG-11   & 0.61 & \textbf{50.64} &  29.20\\
R-PMG-20 & 1.20 & \textbf{55.61} &  25.59\\
\hline
\end{tabular}

%% file: include/nmg/tab_imagenet.tex
\begin{tabular}{@{}l|r|r|rr|rr@{}}
                                 & Params          & FLOPs           & \multicolumn{2}{c|}{val, 10-crop} & \multicolumn{2}{c}{val, single-crop}\\
Method                           & ($\times 10^6$) & ($\times 10^9$) & Top-1          & Top-5            & Top-1          & Top-5\\
\Hline
VGG-16~\cite{vgg16}              & 138.0           & 15.47           & 28.07          & 9.33             & \dash          & \dash \\
ResNet-34 C~\cite{he2015deep}    & 21.8            &  3.66           & 24.19          & 7.40             & \dash          & \dash \\
ResNet-50~\cite{he2015deep}      & 25.6            &  4.46           & 22.85          & 6.71             & \dash          & \dash \\
WRN-34 ($2.0$)
   ~\cite{wideresnet}            & 48.6            & 14.09           & \dash          & \dash            & \textbf{24.50} & 7.58 \\
R-MG-34                          & 32.9            &  5.76           & \textbf{22.42} & \textbf{6.12}    & \textbf{24.51} & \textbf{7.46} \\
R-PMG-30-SEG                     & 31.9            &  2.77           & 23.60          & 6.89             & 26.50          & 8.63\\
\hline
\end{tabular}

%% file: include/nmg/fig_seg_content.tex
\setlength\fboxsep{0pt}
\begin{minipage}[t]{0.49\linewidth}
   \vspace{0pt}
   \begin{center}
   \begin{minipage}[t]{0.09\linewidth}
      \vspace{0pt}
      \begin{center}
         \tiny{\textbf{\textsf{Input}}}\\
         \fbox{\includegraphics[width=1.0\linewidth]{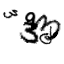}}
      \end{center}
   \end{minipage}
   \hfill
   \begin{minipage}[t]{0.90\linewidth}
      \vspace{0pt}
      \begin{center}
         \begin{minipage}[t]{0.94\linewidth}
            \vspace{0pt}
            \begin{center}
            \begin{minipage}[t]{0.09\linewidth}
               \vspace{0pt}
               \begin{center}
               \tiny{\textbf{\textsf{Digit: 0}}}\\
               \fbox{\includegraphics[width=1.0\linewidth]{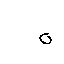}}
               \fbox{\includegraphics[width=1.0\linewidth]{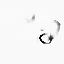}}
               \fbox{\includegraphics[width=1.0\linewidth]{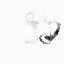}}
               \fbox{\includegraphics[width=1.0\linewidth]{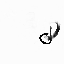}}
               \fbox{\includegraphics[width=1.0\linewidth]{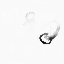}}
               \fbox{\includegraphics[width=1.0\linewidth]{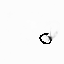}}
               \end{center}
            \end{minipage}
            \begin{minipage}[t]{0.09\linewidth}
               \vspace{0pt}
               \begin{center}
               \tiny{\textbf{\textsf{\phantom{g}1\phantom{g}}}}\\
               \fbox{\includegraphics[width=1.0\linewidth]{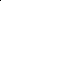}}
               \fbox{\includegraphics[width=1.0\linewidth]{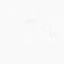}}
               \fbox{\includegraphics[width=1.0\linewidth]{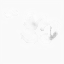}}
               \fbox{\includegraphics[width=1.0\linewidth]{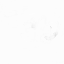}}
               \fbox{\includegraphics[width=1.0\linewidth]{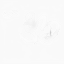}}
               \fbox{\includegraphics[width=1.0\linewidth]{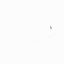}}
               \end{center}
            \end{minipage}
            \begin{minipage}[t]{0.09\linewidth}
               \vspace{0pt}
               \begin{center}
               \tiny{\textbf{\textsf{\phantom{g}2\phantom{g}}}}\\
               \fbox{\includegraphics[width=1.0\linewidth]{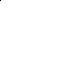}}
               \fbox{\includegraphics[width=1.0\linewidth]{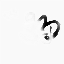}}
               \fbox{\includegraphics[width=1.0\linewidth]{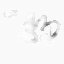}}
               \fbox{\includegraphics[width=1.0\linewidth]{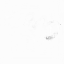}}
               \fbox{\includegraphics[width=1.0\linewidth]{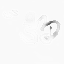}}
               \fbox{\includegraphics[width=1.0\linewidth]{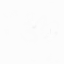}}
               \end{center}
            \end{minipage}
            \begin{minipage}[t]{0.09\linewidth}
               \vspace{0pt}
               \begin{center}
               \tiny{\textbf{\textsf{\phantom{g}3\phantom{g}}}}\\
               \fbox{\includegraphics[width=1.0\linewidth]{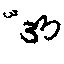}}
               \fbox{\includegraphics[width=1.0\linewidth]{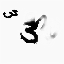}}
               \fbox{\includegraphics[width=1.0\linewidth]{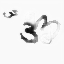}}
               \fbox{\includegraphics[width=1.0\linewidth]{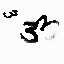}}
               \fbox{\includegraphics[width=1.0\linewidth]{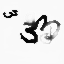}}
               \fbox{\includegraphics[width=1.0\linewidth]{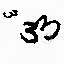}}
               \end{center}
            \end{minipage}
            \begin{minipage}[t]{0.09\linewidth}
               \vspace{0pt}
               \begin{center}
               \tiny{\textbf{\textsf{\phantom{g}4\phantom{g}}}}\\
               \fbox{\includegraphics[width=1.0\linewidth]{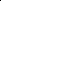}}
               \fbox{\includegraphics[width=1.0\linewidth]{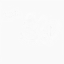}}
               \fbox{\includegraphics[width=1.0\linewidth]{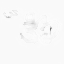}}
               \fbox{\includegraphics[width=1.0\linewidth]{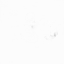}}
               \fbox{\includegraphics[width=1.0\linewidth]{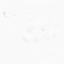}}
               \fbox{\includegraphics[width=1.0\linewidth]{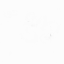}}
               \end{center}
            \end{minipage}
            \begin{minipage}[t]{0.09\linewidth}
               \vspace{0pt}
               \begin{center}
               \tiny{\textbf{\textsf{\phantom{g}5\phantom{g}}}}\\
               \fbox{\includegraphics[width=1.0\linewidth]{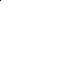}}
               \fbox{\includegraphics[width=1.0\linewidth]{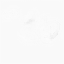}}
               \fbox{\includegraphics[width=1.0\linewidth]{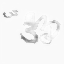}}
               \fbox{\includegraphics[width=1.0\linewidth]{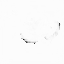}}
               \fbox{\includegraphics[width=1.0\linewidth]{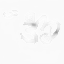}}
               \fbox{\includegraphics[width=1.0\linewidth]{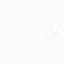}}
               \end{center}
            \end{minipage}
            \begin{minipage}[t]{0.09\linewidth}
               \vspace{0pt}
               \begin{center}
               \tiny{\textbf{\textsf{\phantom{g}6\phantom{g}}}}\\
               \fbox{\includegraphics[width=1.0\linewidth]{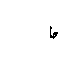}}
               \fbox{\includegraphics[width=1.0\linewidth]{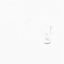}}
               \fbox{\includegraphics[width=1.0\linewidth]{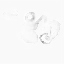}}
               \fbox{\includegraphics[width=1.0\linewidth]{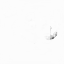}}
               \fbox{\includegraphics[width=1.0\linewidth]{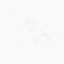}}
               \fbox{\includegraphics[width=1.0\linewidth]{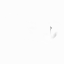}}
               \end{center}
            \end{minipage}
            \begin{minipage}[t]{0.09\linewidth}
               \vspace{0pt}
               \begin{center}
               \tiny{\textbf{\textsf{\phantom{g}7\phantom{g}}}}\\
               \fbox{\includegraphics[width=1.0\linewidth]{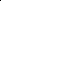}}
               \fbox{\includegraphics[width=1.0\linewidth]{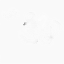}}
               \fbox{\includegraphics[width=1.0\linewidth]{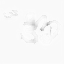}}
               \fbox{\includegraphics[width=1.0\linewidth]{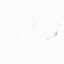}}
               \fbox{\includegraphics[width=1.0\linewidth]{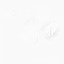}}
               \fbox{\includegraphics[width=1.0\linewidth]{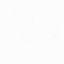}}
               \end{center}
            \end{minipage}
            \begin{minipage}[t]{0.09\linewidth}
               \vspace{0pt}
               \begin{center}
               \tiny{\textbf{\textsf{\phantom{g}8\phantom{g}}}}\\
               \fbox{\includegraphics[width=1.0\linewidth]{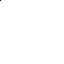}}
               \fbox{\includegraphics[width=1.0\linewidth]{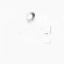}}
               \fbox{\includegraphics[width=1.0\linewidth]{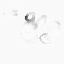}}
               \fbox{\includegraphics[width=1.0\linewidth]{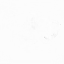}}
               \fbox{\includegraphics[width=1.0\linewidth]{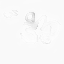}}
               \fbox{\includegraphics[width=1.0\linewidth]{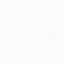}}
               \end{center}
            \end{minipage}
            \begin{minipage}[t]{0.09\linewidth}
               \vspace{0pt}
               \begin{center}
               \tiny{\textbf{\textsf{\phantom{g}9\phantom{g}}}}\\
               \fbox{\includegraphics[width=1.0\linewidth]{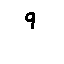}}
               \fbox{\includegraphics[width=1.0\linewidth]{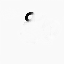}}
               \fbox{\includegraphics[width=1.0\linewidth]{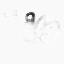}}
               \fbox{\includegraphics[width=1.0\linewidth]{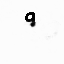}}
               \fbox{\includegraphics[width=1.0\linewidth]{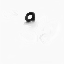}}
               \fbox{\includegraphics[width=1.0\linewidth]{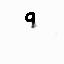}}
               \end{center}
            \end{minipage}
            \end{center}
         \end{minipage}
         \begin{minipage}[t]{0.04\linewidth}
            \vspace{0pt}
            \begin{rotate}{270}
               \tiny{\textbf{\textsf{
                  \hspace*{8pt}
                  True
                  \hspace*{4pt}
                  U-NET
                  \hspace*{6pt}
                  SG
                  \hspace*{8pt}
                  R-SG
                  \hspace*{5pt}
                  PMG
                  \hspace*{4pt}
                  R-PMG
               }}}
            \end{rotate}
         \end{minipage}
      \end{center}
   \end{minipage}
   \end{center}
\end{minipage}
\hfill
\begin{minipage}[t]{0.49\linewidth}
   \vspace{0pt}
   \begin{center}
   \begin{minipage}[t]{0.09\linewidth}
      \vspace{0pt}
      \begin{center}
         \tiny{\textbf{\textsf{Input}}}\\
         \fbox{\includegraphics[width=1.0\linewidth]{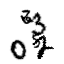}}
      \end{center}
   \end{minipage}
   \hfill
   \begin{minipage}[t]{0.90\linewidth}
      \vspace{0pt}
      \begin{center}
         \begin{minipage}[t]{0.94\linewidth}
            \vspace{0pt}
            \begin{center}
            \begin{minipage}[t]{0.09\linewidth}
               \vspace{0pt}
               \begin{center}
               \tiny{\textbf{\textsf{Digit: 0}}}\\
               \fbox{\includegraphics[width=1.0\linewidth]{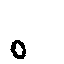}}
               \fbox{\includegraphics[width=1.0\linewidth]{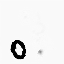}}
               \fbox{\includegraphics[width=1.0\linewidth]{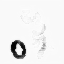}}
               \fbox{\includegraphics[width=1.0\linewidth]{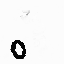}}
               \fbox{\includegraphics[width=1.0\linewidth]{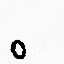}}
               \fbox{\includegraphics[width=1.0\linewidth]{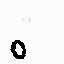}}
               \end{center}
            \end{minipage}
            \begin{minipage}[t]{0.09\linewidth}
               \vspace{0pt}
               \begin{center}
               \tiny{\textbf{\textsf{\phantom{g}1\phantom{g}}}}\\
               \fbox{\includegraphics[width=1.0\linewidth]{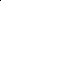}}
               \fbox{\includegraphics[width=1.0\linewidth]{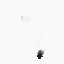}}
               \fbox{\includegraphics[width=1.0\linewidth]{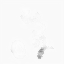}}
               \fbox{\includegraphics[width=1.0\linewidth]{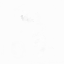}}
               \fbox{\includegraphics[width=1.0\linewidth]{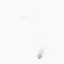}}
               \fbox{\includegraphics[width=1.0\linewidth]{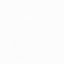}}
               \end{center}
            \end{minipage}
            \begin{minipage}[t]{0.09\linewidth}
               \vspace{0pt}
               \begin{center}
               \tiny{\textbf{\textsf{\phantom{g}2\phantom{g}}}}\\
               \fbox{\includegraphics[width=1.0\linewidth]{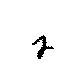}}
               \fbox{\includegraphics[width=1.0\linewidth]{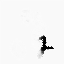}}
               \fbox{\includegraphics[width=1.0\linewidth]{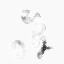}}
               \fbox{\includegraphics[width=1.0\linewidth]{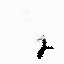}}
               \fbox{\includegraphics[width=1.0\linewidth]{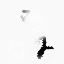}}
               \fbox{\includegraphics[width=1.0\linewidth]{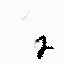}}
               \end{center}
            \end{minipage}
            \begin{minipage}[t]{0.09\linewidth}
               \vspace{0pt}
               \begin{center}
               \tiny{\textbf{\textsf{\phantom{g}3\phantom{g}}}}\\
               \fbox{\includegraphics[width=1.0\linewidth]{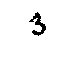}}
               \fbox{\includegraphics[width=1.0\linewidth]{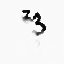}}
               \fbox{\includegraphics[width=1.0\linewidth]{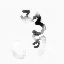}}
               \fbox{\includegraphics[width=1.0\linewidth]{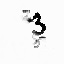}}
               \fbox{\includegraphics[width=1.0\linewidth]{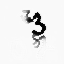}}
               \fbox{\includegraphics[width=1.0\linewidth]{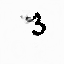}}
               \end{center}
            \end{minipage}
            \begin{minipage}[t]{0.09\linewidth}
               \vspace{0pt}
               \begin{center}
               \tiny{\textbf{\textsf{\phantom{g}4\phantom{g}}}}\\
               \fbox{\includegraphics[width=1.0\linewidth]{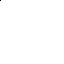}}
               \fbox{\includegraphics[width=1.0\linewidth]{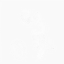}}
               \fbox{\includegraphics[width=1.0\linewidth]{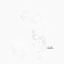}}
               \fbox{\includegraphics[width=1.0\linewidth]{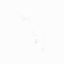}}
               \fbox{\includegraphics[width=1.0\linewidth]{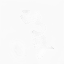}}
               \fbox{\includegraphics[width=1.0\linewidth]{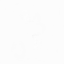}}
               \end{center}
            \end{minipage}
            \begin{minipage}[t]{0.09\linewidth}
               \vspace{0pt}
               \begin{center}
               \tiny{\textbf{\textsf{\phantom{g}5\phantom{g}}}}\\
               \fbox{\includegraphics[width=1.0\linewidth]{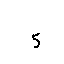}}
               \fbox{\includegraphics[width=1.0\linewidth]{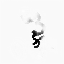}}
               \fbox{\includegraphics[width=1.0\linewidth]{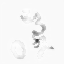}}
               \fbox{\includegraphics[width=1.0\linewidth]{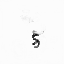}}
               \fbox{\includegraphics[width=1.0\linewidth]{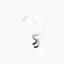}}
               \fbox{\includegraphics[width=1.0\linewidth]{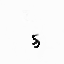}}
               \end{center}
            \end{minipage}
            \begin{minipage}[t]{0.09\linewidth}
               \vspace{0pt}
               \begin{center}
               \tiny{\textbf{\textsf{\phantom{g}6\phantom{g}}}}\\
               \fbox{\includegraphics[width=1.0\linewidth]{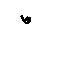}}
               \fbox{\includegraphics[width=1.0\linewidth]{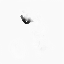}}
               \fbox{\includegraphics[width=1.0\linewidth]{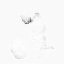}}
               \fbox{\includegraphics[width=1.0\linewidth]{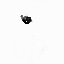}}
               \fbox{\includegraphics[width=1.0\linewidth]{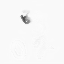}}
               \fbox{\includegraphics[width=1.0\linewidth]{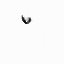}}
               \end{center}
            \end{minipage}
            \begin{minipage}[t]{0.09\linewidth}
               \vspace{0pt}
               \begin{center}
               \tiny{\textbf{\textsf{\phantom{g}7\phantom{g}}}}\\
               \fbox{\includegraphics[width=1.0\linewidth]{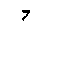}}
               \fbox{\includegraphics[width=1.0\linewidth]{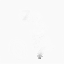}}
               \fbox{\includegraphics[width=1.0\linewidth]{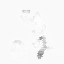}}
               \fbox{\includegraphics[width=1.0\linewidth]{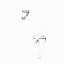}}
               \fbox{\includegraphics[width=1.0\linewidth]{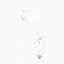}}
               \fbox{\includegraphics[width=1.0\linewidth]{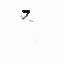}}
               \end{center}
            \end{minipage}
            \begin{minipage}[t]{0.09\linewidth}
               \vspace{0pt}
               \begin{center}
               \tiny{\textbf{\textsf{\phantom{g}8\phantom{g}}}}\\
               \fbox{\includegraphics[width=1.0\linewidth]{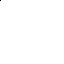}}
               \fbox{\includegraphics[width=1.0\linewidth]{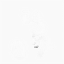}}
               \fbox{\includegraphics[width=1.0\linewidth]{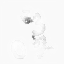}}
               \fbox{\includegraphics[width=1.0\linewidth]{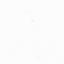}}
               \fbox{\includegraphics[width=1.0\linewidth]{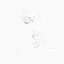}}
               \fbox{\includegraphics[width=1.0\linewidth]{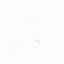}}
               \end{center}
            \end{minipage}
            \begin{minipage}[t]{0.09\linewidth}
               \vspace{0pt}
               \begin{center}
               \tiny{\textbf{\textsf{\phantom{g}9\phantom{g}}}}\\
               \fbox{\includegraphics[width=1.0\linewidth]{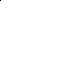}}
               \fbox{\includegraphics[width=1.0\linewidth]{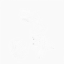}}
               \fbox{\includegraphics[width=1.0\linewidth]{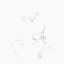}}
               \fbox{\includegraphics[width=1.0\linewidth]{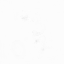}}
               \fbox{\includegraphics[width=1.0\linewidth]{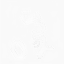}}
               \fbox{\includegraphics[width=1.0\linewidth]{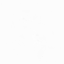}}
               \end{center}
            \end{minipage}
            \end{center}
         \end{minipage}
         \begin{minipage}[t]{0.04\linewidth}
            \vspace{0pt}
            \begin{rotate}{270}
               \tiny{\textbf{\textsf{
                  \hspace*{8pt}
                  True
                  \hspace*{4pt}
                  U-NET
                  \hspace*{6pt}
                  SG
                  \hspace*{8pt}
                  R-SG
                  \hspace*{5pt}
                  PMG
                  \hspace*{4pt}
                  R-PMG
               }}}
            \end{rotate}
         \end{minipage}
      \end{center}
   \end{minipage}
   \end{center}
\end{minipage}\\
\vspace{0.02\linewidth}
\setlength\fboxsep{0pt}
\begin{minipage}[t]{0.49\linewidth}
   \vspace{0pt}
   \begin{center}
   \begin{minipage}[t]{0.09\linewidth}
      \vspace{0pt}
      \begin{center}
         \tiny{\textbf{\textsf{Input}}}\\
         \fbox{\includegraphics[width=1.0\linewidth]{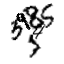}}
      \end{center}
   \end{minipage}
   \hfill
   \begin{minipage}[t]{0.90\linewidth}
      \vspace{0pt}
      \begin{center}
         \begin{minipage}[t]{0.94\linewidth}
            \vspace{0pt}
            \begin{center}
            \begin{minipage}[t]{0.09\linewidth}
               \vspace{0pt}
               \begin{center}
               \tiny{\textbf{\textsf{Digit: 0}}}\\
               \fbox{\includegraphics[width=1.0\linewidth]{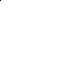}}
               \fbox{\includegraphics[width=1.0\linewidth]{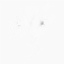}}
               \fbox{\includegraphics[width=1.0\linewidth]{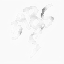}}
               \fbox{\includegraphics[width=1.0\linewidth]{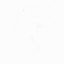}}
               \fbox{\includegraphics[width=1.0\linewidth]{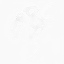}}
               \fbox{\includegraphics[width=1.0\linewidth]{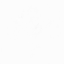}}
               \end{center}
            \end{minipage}
            \begin{minipage}[t]{0.09\linewidth}
               \vspace{0pt}
               \begin{center}
               \tiny{\textbf{\textsf{\phantom{g}1\phantom{g}}}}\\
               \fbox{\includegraphics[width=1.0\linewidth]{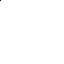}}
               \fbox{\includegraphics[width=1.0\linewidth]{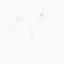}}
               \fbox{\includegraphics[width=1.0\linewidth]{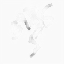}}
               \fbox{\includegraphics[width=1.0\linewidth]{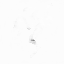}}
               \fbox{\includegraphics[width=1.0\linewidth]{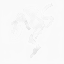}}
               \fbox{\includegraphics[width=1.0\linewidth]{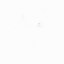}}
               \end{center}
            \end{minipage}
            \begin{minipage}[t]{0.09\linewidth}
               \vspace{0pt}
               \begin{center}
               \tiny{\textbf{\textsf{\phantom{g}2\phantom{g}}}}\\
               \fbox{\includegraphics[width=1.0\linewidth]{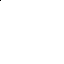}}
               \fbox{\includegraphics[width=1.0\linewidth]{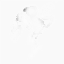}}
               \fbox{\includegraphics[width=1.0\linewidth]{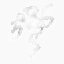}}
               \fbox{\includegraphics[width=1.0\linewidth]{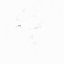}}
               \fbox{\includegraphics[width=1.0\linewidth]{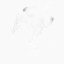}}
               \fbox{\includegraphics[width=1.0\linewidth]{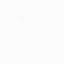}}
               \end{center}
            \end{minipage}
            \begin{minipage}[t]{0.09\linewidth}
               \vspace{0pt}
               \begin{center}
               \tiny{\textbf{\textsf{\phantom{g}3\phantom{g}}}}\\
               \fbox{\includegraphics[width=1.0\linewidth]{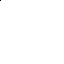}}
               \fbox{\includegraphics[width=1.0\linewidth]{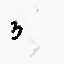}}
               \fbox{\includegraphics[width=1.0\linewidth]{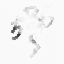}}
               \fbox{\includegraphics[width=1.0\linewidth]{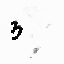}}
               \fbox{\includegraphics[width=1.0\linewidth]{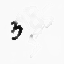}}
               \fbox{\includegraphics[width=1.0\linewidth]{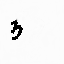}}
               \end{center}
            \end{minipage}
            \begin{minipage}[t]{0.09\linewidth}
               \vspace{0pt}
               \begin{center}
               \tiny{\textbf{\textsf{\phantom{g}4\phantom{g}}}}\\
               \fbox{\includegraphics[width=1.0\linewidth]{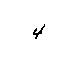}}
               \fbox{\includegraphics[width=1.0\linewidth]{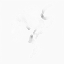}}
               \fbox{\includegraphics[width=1.0\linewidth]{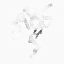}}
               \fbox{\includegraphics[width=1.0\linewidth]{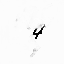}}
               \fbox{\includegraphics[width=1.0\linewidth]{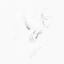}}
               \fbox{\includegraphics[width=1.0\linewidth]{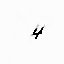}}
               \end{center}
            \end{minipage}
            \begin{minipage}[t]{0.09\linewidth}
               \vspace{0pt}
               \begin{center}
               \tiny{\textbf{\textsf{\phantom{g}5\phantom{g}}}}\\
               \fbox{\includegraphics[width=1.0\linewidth]{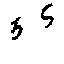}}
               \fbox{\includegraphics[width=1.0\linewidth]{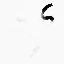}}
               \fbox{\includegraphics[width=1.0\linewidth]{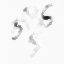}}
               \fbox{\includegraphics[width=1.0\linewidth]{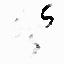}}
               \fbox{\includegraphics[width=1.0\linewidth]{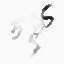}}
               \fbox{\includegraphics[width=1.0\linewidth]{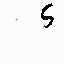}}
               \end{center}
            \end{minipage}
            \begin{minipage}[t]{0.09\linewidth}
               \vspace{0pt}
               \begin{center}
               \tiny{\textbf{\textsf{\phantom{g}6\phantom{g}}}}\\
               \fbox{\includegraphics[width=1.0\linewidth]{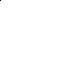}}
               \fbox{\includegraphics[width=1.0\linewidth]{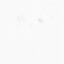}}
               \fbox{\includegraphics[width=1.0\linewidth]{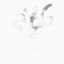}}
               \fbox{\includegraphics[width=1.0\linewidth]{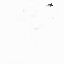}}
               \fbox{\includegraphics[width=1.0\linewidth]{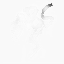}}
               \fbox{\includegraphics[width=1.0\linewidth]{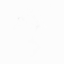}}
               \end{center}
            \end{minipage}
            \begin{minipage}[t]{0.09\linewidth}
               \vspace{0pt}
               \begin{center}
               \tiny{\textbf{\textsf{\phantom{g}7\phantom{g}}}}\\
               \fbox{\includegraphics[width=1.0\linewidth]{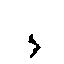}}
               \fbox{\includegraphics[width=1.0\linewidth]{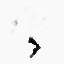}}
               \fbox{\includegraphics[width=1.0\linewidth]{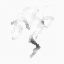}}
               \fbox{\includegraphics[width=1.0\linewidth]{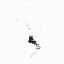}}
               \fbox{\includegraphics[width=1.0\linewidth]{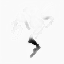}}
               \fbox{\includegraphics[width=1.0\linewidth]{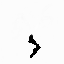}}
               \end{center}
            \end{minipage}
            \begin{minipage}[t]{0.09\linewidth}
               \vspace{0pt}
               \begin{center}
               \tiny{\textbf{\textsf{\phantom{g}8\phantom{g}}}}\\
               \fbox{\includegraphics[width=1.0\linewidth]{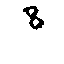}}
               \fbox{\includegraphics[width=1.0\linewidth]{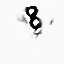}}
               \fbox{\includegraphics[width=1.0\linewidth]{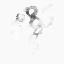}}
               \fbox{\includegraphics[width=1.0\linewidth]{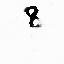}}
               \fbox{\includegraphics[width=1.0\linewidth]{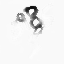}}
               \fbox{\includegraphics[width=1.0\linewidth]{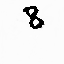}}
               \end{center}
            \end{minipage}
            \begin{minipage}[t]{0.09\linewidth}
               \vspace{0pt}
               \begin{center}
               \tiny{\textbf{\textsf{\phantom{g}9\phantom{g}}}}\\
               \fbox{\includegraphics[width=1.0\linewidth]{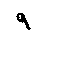}}
               \fbox{\includegraphics[width=1.0\linewidth]{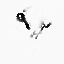}}
               \fbox{\includegraphics[width=1.0\linewidth]{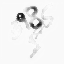}}
               \fbox{\includegraphics[width=1.0\linewidth]{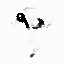}}
               \fbox{\includegraphics[width=1.0\linewidth]{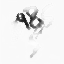}}
               \fbox{\includegraphics[width=1.0\linewidth]{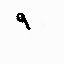}}
               \end{center}
            \end{minipage}
            \end{center}
         \end{minipage}
         \begin{minipage}[t]{0.04\linewidth}
            \vspace{0pt}
            \begin{rotate}{270}
               \tiny{\textbf{\textsf{
                  \hspace*{8pt}
                  True
                  \hspace*{4pt}
                  U-NET
                  \hspace*{6pt}
                  SG
                  \hspace*{8pt}
                  R-SG
                  \hspace*{5pt}
                  PMG
                  \hspace*{4pt}
                  R-PMG
               }}}
            \end{rotate}
         \end{minipage}
      \end{center}
   \end{minipage}
   \end{center}
\end{minipage}
\hfill
\begin{minipage}[t]{0.49\linewidth}
   \vspace{0pt}
   \begin{center}
   \begin{minipage}[t]{0.09\linewidth}
      \vspace{0pt}
      \begin{center}
         \tiny{\textbf{\textsf{Input}}}\\
         \fbox{\includegraphics[width=1.0\linewidth]{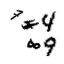}}
      \end{center}
   \end{minipage}
   \hfill
   \begin{minipage}[t]{0.90\linewidth}
      \vspace{0pt}
      \begin{center}
         \begin{minipage}[t]{0.94\linewidth}
            \vspace{0pt}
            \begin{center}
            \begin{minipage}[t]{0.09\linewidth}
               \vspace{0pt}
               \begin{center}
               \tiny{\textbf{\textsf{Digit: 0}}}\\
               \fbox{\includegraphics[width=1.0\linewidth]{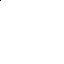}}
               \fbox{\includegraphics[width=1.0\linewidth]{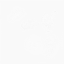}}
               \fbox{\includegraphics[width=1.0\linewidth]{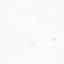}}
               \fbox{\includegraphics[width=1.0\linewidth]{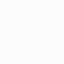}}
               \fbox{\includegraphics[width=1.0\linewidth]{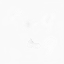}}
               \fbox{\includegraphics[width=1.0\linewidth]{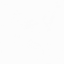}}
               \end{center}
            \end{minipage}
            \begin{minipage}[t]{0.09\linewidth}
               \vspace{0pt}
               \begin{center}
               \tiny{\textbf{\textsf{\phantom{g}1\phantom{g}}}}\\
               \fbox{\includegraphics[width=1.0\linewidth]{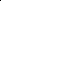}}
               \fbox{\includegraphics[width=1.0\linewidth]{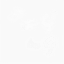}}
               \fbox{\includegraphics[width=1.0\linewidth]{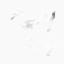}}
               \fbox{\includegraphics[width=1.0\linewidth]{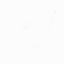}}
               \fbox{\includegraphics[width=1.0\linewidth]{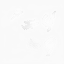}}
               \fbox{\includegraphics[width=1.0\linewidth]{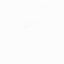}}
               \end{center}
            \end{minipage}
            \begin{minipage}[t]{0.09\linewidth}
               \vspace{0pt}
               \begin{center}
               \tiny{\textbf{\textsf{\phantom{g}2\phantom{g}}}}\\
               \fbox{\includegraphics[width=1.0\linewidth]{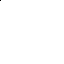}}
               \fbox{\includegraphics[width=1.0\linewidth]{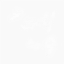}}
               \fbox{\includegraphics[width=1.0\linewidth]{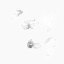}}
               \fbox{\includegraphics[width=1.0\linewidth]{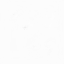}}
               \fbox{\includegraphics[width=1.0\linewidth]{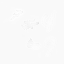}}
               \fbox{\includegraphics[width=1.0\linewidth]{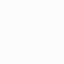}}
               \end{center}
            \end{minipage}
            \begin{minipage}[t]{0.09\linewidth}
               \vspace{0pt}
               \begin{center}
               \tiny{\textbf{\textsf{\phantom{g}3\phantom{g}}}}\\
               \fbox{\includegraphics[width=1.0\linewidth]{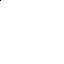}}
               \fbox{\includegraphics[width=1.0\linewidth]{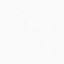}}
               \fbox{\includegraphics[width=1.0\linewidth]{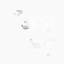}}
               \fbox{\includegraphics[width=1.0\linewidth]{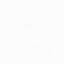}}
               \fbox{\includegraphics[width=1.0\linewidth]{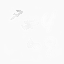}}
               \fbox{\includegraphics[width=1.0\linewidth]{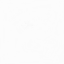}}
               \end{center}
            \end{minipage}
            \begin{minipage}[t]{0.09\linewidth}
               \vspace{0pt}
               \begin{center}
               \tiny{\textbf{\textsf{\phantom{g}4\phantom{g}}}}\\
               \fbox{\includegraphics[width=1.0\linewidth]{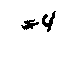}}
               \fbox{\includegraphics[width=1.0\linewidth]{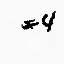}}
               \fbox{\includegraphics[width=1.0\linewidth]{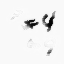}}
               \fbox{\includegraphics[width=1.0\linewidth]{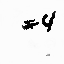}}
               \fbox{\includegraphics[width=1.0\linewidth]{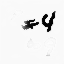}}
               \fbox{\includegraphics[width=1.0\linewidth]{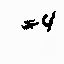}}
               \end{center}
            \end{minipage}
            \begin{minipage}[t]{0.09\linewidth}
               \vspace{0pt}
               \begin{center}
               \tiny{\textbf{\textsf{\phantom{g}5\phantom{g}}}}\\
               \fbox{\includegraphics[width=1.0\linewidth]{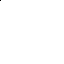}}
               \fbox{\includegraphics[width=1.0\linewidth]{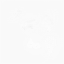}}
               \fbox{\includegraphics[width=1.0\linewidth]{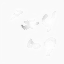}}
               \fbox{\includegraphics[width=1.0\linewidth]{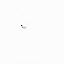}}
               \fbox{\includegraphics[width=1.0\linewidth]{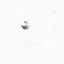}}
               \fbox{\includegraphics[width=1.0\linewidth]{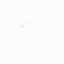}}
               \end{center}
            \end{minipage}
            \begin{minipage}[t]{0.09\linewidth}
               \vspace{0pt}
               \begin{center}
               \tiny{\textbf{\textsf{\phantom{g}6\phantom{g}}}}\\
               \fbox{\includegraphics[width=1.0\linewidth]{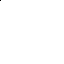}}
               \fbox{\includegraphics[width=1.0\linewidth]{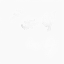}}
               \fbox{\includegraphics[width=1.0\linewidth]{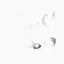}}
               \fbox{\includegraphics[width=1.0\linewidth]{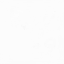}}
               \fbox{\includegraphics[width=1.0\linewidth]{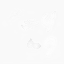}}
               \fbox{\includegraphics[width=1.0\linewidth]{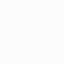}}
               \end{center}
            \end{minipage}
            \begin{minipage}[t]{0.09\linewidth}
               \vspace{0pt}
               \begin{center}
               \tiny{\textbf{\textsf{\phantom{g}7\phantom{g}}}}\\
               \fbox{\includegraphics[width=1.0\linewidth]{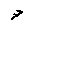}}
               \fbox{\includegraphics[width=1.0\linewidth]{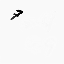}}
               \fbox{\includegraphics[width=1.0\linewidth]{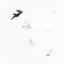}}
               \fbox{\includegraphics[width=1.0\linewidth]{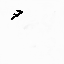}}
               \fbox{\includegraphics[width=1.0\linewidth]{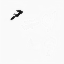}}
               \fbox{\includegraphics[width=1.0\linewidth]{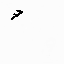}}
               \end{center}
            \end{minipage}
            \begin{minipage}[t]{0.09\linewidth}
               \vspace{0pt}
               \begin{center}
               \tiny{\textbf{\textsf{\phantom{g}8\phantom{g}}}}\\
               \fbox{\includegraphics[width=1.0\linewidth]{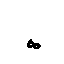}}
               \fbox{\includegraphics[width=1.0\linewidth]{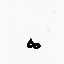}}
               \fbox{\includegraphics[width=1.0\linewidth]{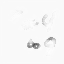}}
               \fbox{\includegraphics[width=1.0\linewidth]{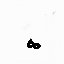}}
               \fbox{\includegraphics[width=1.0\linewidth]{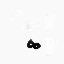}}
               \fbox{\includegraphics[width=1.0\linewidth]{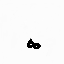}}
               \end{center}
            \end{minipage}
            \begin{minipage}[t]{0.09\linewidth}
               \vspace{0pt}
               \begin{center}
               \tiny{\textbf{\textsf{\phantom{g}9\phantom{g}}}}\\
               \fbox{\includegraphics[width=1.0\linewidth]{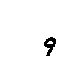}}
               \fbox{\includegraphics[width=1.0\linewidth]{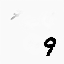}}
               \fbox{\includegraphics[width=1.0\linewidth]{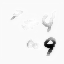}}
               \fbox{\includegraphics[width=1.0\linewidth]{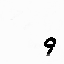}}
               \fbox{\includegraphics[width=1.0\linewidth]{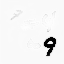}}
               \fbox{\includegraphics[width=1.0\linewidth]{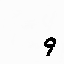}}
               \end{center}
            \end{minipage}
            \end{center}
         \end{minipage}
         \begin{minipage}[t]{0.04\linewidth}
            \vspace{0pt}
            \begin{rotate}{270}
               \tiny{\textbf{\textsf{
                  \hspace*{8pt}
                  True
                  \hspace*{4pt}
                  U-NET
                  \hspace*{6pt}
                  SG
                  \hspace*{8pt}
                  R-SG
                  \hspace*{5pt}
                  PMG
                  \hspace*{4pt}
                  R-PMG
               }}}
            \end{rotate}
         \end{minipage}
      \end{center}
   \end{minipage}
   \end{center}
\end{minipage}

%% file: include/nmg/fig_spt_content.tex
\setlength\fboxsep{0pt}
\begin{minipage}[t]{0.54\linewidth}
   \vspace{0pt}
   \begin{center}
   \begin{minipage}[t]{0.13143\linewidth}
      \vspace{0pt}
      \begin{center}
      \tiny{\textbf{\textsf{Input}}}\\
      \fbox{\includegraphics[width=1.0\linewidth]{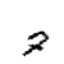}}
      \fbox{\includegraphics[width=1.0\linewidth]{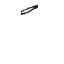}}
      \fbox{\includegraphics[width=1.0\linewidth]{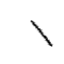}}
      \fbox{\includegraphics[width=1.0\linewidth]{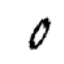}}
      \fbox{\includegraphics[width=1.0\linewidth]{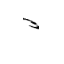}}
      \fbox{\includegraphics[width=1.0\linewidth]{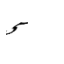}}
      \fbox{\includegraphics[width=1.0\linewidth]{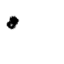}}
      \end{center}
   \end{minipage}
   \begin{minipage}[t]{0.13143\linewidth}
      \vspace{0pt}
      \begin{center}
      \tiny{\textbf{\textsf{\phantom{p}True\phantom{p}}}}\\
      \fbox{\includegraphics[width=1.0\linewidth]{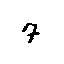}}
      \fbox{\includegraphics[width=1.0\linewidth]{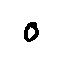}}
      \fbox{\includegraphics[width=1.0\linewidth]{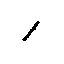}}
      \fbox{\includegraphics[width=1.0\linewidth]{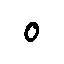}}
      \fbox{\includegraphics[width=1.0\linewidth]{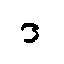}}
      \fbox{\includegraphics[width=1.0\linewidth]{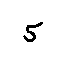}}
      \fbox{\includegraphics[width=1.0\linewidth]{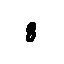}}
      \end{center}
   \end{minipage}
   \begin{minipage}[t]{0.13143\linewidth}
      \vspace{0pt}
      \begin{center}
      \tiny{\textbf{\textsf{\phantom{p}U-NET\phantom{p}}}}\\
      \fbox{\includegraphics[width=1.0\linewidth]{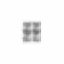}}
      \fbox{\includegraphics[width=1.0\linewidth]{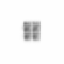}}
      \fbox{\includegraphics[width=1.0\linewidth]{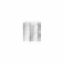}}
      \fbox{\includegraphics[width=1.0\linewidth]{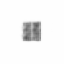}}
      \fbox{\includegraphics[width=1.0\linewidth]{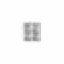}}
      \fbox{\includegraphics[width=1.0\linewidth]{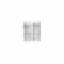}}
      \fbox{\includegraphics[width=1.0\linewidth]{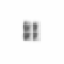}}
      \end{center}
   \end{minipage}
   \begin{minipage}[t]{0.13143\linewidth}
      \vspace{0pt}
      \begin{center}
      \tiny{\textbf{\textsf{\phantom{p}SG}\phantom{p}}}\\
      \fbox{\includegraphics[width=1.0\linewidth]{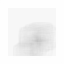}}
      \fbox{\includegraphics[width=1.0\linewidth]{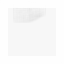}}
      \fbox{\includegraphics[width=1.0\linewidth]{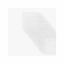}}
      \fbox{\includegraphics[width=1.0\linewidth]{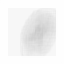}}
      \fbox{\includegraphics[width=1.0\linewidth]{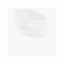}}
      \fbox{\includegraphics[width=1.0\linewidth]{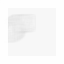}}
      \fbox{\includegraphics[width=1.0\linewidth]{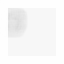}}
      \end{center}
   \end{minipage}
   \begin{minipage}[t]{0.13143\linewidth}
      \vspace{0pt}
      \begin{center}
      \tiny{\textbf{\textsf{\phantom{p}R-SG\phantom{p}}}}\\
      \fbox{\includegraphics[width=1.0\linewidth]{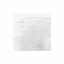}}
      \fbox{\includegraphics[width=1.0\linewidth]{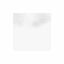}}
      \fbox{\includegraphics[width=1.0\linewidth]{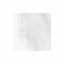}}
      \fbox{\includegraphics[width=1.0\linewidth]{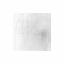}}
      \fbox{\includegraphics[width=1.0\linewidth]{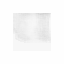}}
      \fbox{\includegraphics[width=1.0\linewidth]{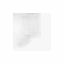}}
      \fbox{\includegraphics[width=1.0\linewidth]{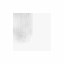}}
      \end{center}
   \end{minipage}
   \begin{minipage}[t]{0.13143\linewidth}
      \vspace{0pt}
      \begin{center}
      \tiny{\textbf{\textsf{\phantom{p}PMG\phantom{p}}}}\\
      \fbox{\includegraphics[width=1.0\linewidth]{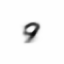}}
      \fbox{\includegraphics[width=1.0\linewidth]{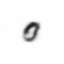}}
      \fbox{\includegraphics[width=1.0\linewidth]{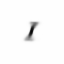}}
      \fbox{\includegraphics[width=1.0\linewidth]{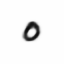}}
      \fbox{\includegraphics[width=1.0\linewidth]{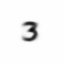}}
      \fbox{\includegraphics[width=1.0\linewidth]{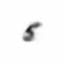}}
      \fbox{\includegraphics[width=1.0\linewidth]{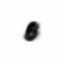}}
      \end{center}
   \end{minipage}
   \begin{minipage}[t]{0.13143\linewidth}
      \vspace{0pt}
      \begin{center}
      \tiny{\textbf{\textsf{\phantom{p}R-PMG\phantom{p}}}}\\
      \fbox{\includegraphics[width=1.0\linewidth]{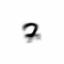}}
      \fbox{\includegraphics[width=1.0\linewidth]{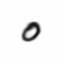}}
      \fbox{\includegraphics[width=1.0\linewidth]{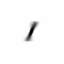}}
      \fbox{\includegraphics[width=1.0\linewidth]{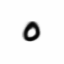}}
      \fbox{\includegraphics[width=1.0\linewidth]{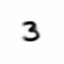}}
      \fbox{\includegraphics[width=1.0\linewidth]{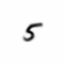}}
      \fbox{\includegraphics[width=1.0\linewidth]{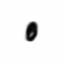}}
      \end{center}
   \end{minipage}\\
   \vspace{0.012\linewidth}
   \scriptsize{\textbf{\textsf{Scaling \& Rotation \& Affine \& Translation}}}
   \end{center}
\end{minipage}
\hfill
\begin{minipage}[t]{0.445\linewidth}
   \vspace{0pt}
   \begin{center}
   \begin{minipage}[t]{0.155\linewidth}
      \vspace{0pt}
      \begin{center}
      \tiny{\textbf{\textsf{Input}}}\\
      \fbox{\includegraphics[width=1.0\linewidth]{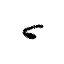}}
      \fbox{\includegraphics[width=1.0\linewidth]{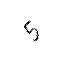}}
      \end{center}
   \end{minipage}
   \begin{minipage}[t]{0.155\linewidth}
      \vspace{0pt}
      \begin{center}
      \tiny{\textbf{\textsf{\phantom{p}True\phantom{p}}}}\\
      \fbox{\includegraphics[width=1.0\linewidth]{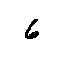}}
      \fbox{\includegraphics[width=1.0\linewidth]{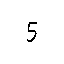}}
      \end{center}
   \end{minipage}
   \begin{minipage}[t]{0.155\linewidth}
      \vspace{0pt}
      \begin{center}
      \tiny{\textbf{\textsf{\phantom{p}U-NET\phantom{p}}}}\\
      \fbox{\includegraphics[width=1.0\linewidth]{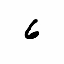}}
      \fbox{\includegraphics[width=1.0\linewidth]{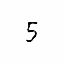}}
      \end{center}
   \end{minipage}
   \hspace{0.0025\linewidth}
   \begin{minipage}[t]{0.155\linewidth}
      \vspace{0pt}
      \begin{center}
      \tiny{\textbf{\textsf{Input}}}\\
      \fbox{\includegraphics[width=1.0\linewidth]{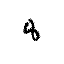}}
      \fbox{\includegraphics[width=1.0\linewidth]{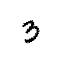}}
      \end{center}
   \end{minipage}
   \begin{minipage}[t]{0.155\linewidth}
      \vspace{0pt}
      \begin{center}
      \tiny{\textbf{\textsf{\phantom{p}True\phantom{p}}}}\\
      \fbox{\includegraphics[width=1.0\linewidth]{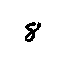}}
      \fbox{\includegraphics[width=1.0\linewidth]{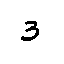}}
      \end{center}
   \end{minipage}
   \begin{minipage}[t]{0.155\linewidth}
      \vspace{0pt}
      \begin{center}
      \tiny{\textbf{\textsf{\phantom{p}U-NET\phantom{p}}}}\\
      \fbox{\includegraphics[width=1.0\linewidth]{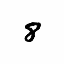}}
      \fbox{\includegraphics[width=1.0\linewidth]{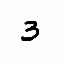}}
      \end{center}
   \end{minipage}\\
   \vspace{0.013\linewidth}
   \scriptsize{\textbf{\textsf{Rotation Only}}}\\
   \begin{minipage}[t]{0.155\linewidth}
      \vspace{0pt}
      \begin{center}
      \tiny{\textbf{\textsf{Input}}}\\
      \fbox{\includegraphics[width=1.0\linewidth]{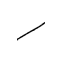}}
      \fbox{\includegraphics[width=1.0\linewidth]{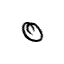}}
      \end{center}
   \end{minipage}
   \begin{minipage}[t]{0.155\linewidth}
      \vspace{0pt}
      \begin{center}
      \tiny{\textbf{\textsf{\phantom{p}True\phantom{p}}}}\\
      \fbox{\includegraphics[width=1.0\linewidth]{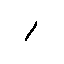}}
      \fbox{\includegraphics[width=1.0\linewidth]{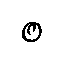}}
      \end{center}
   \end{minipage}
   \begin{minipage}[t]{0.155\linewidth}
      \vspace{0pt}
      \begin{center}
      \tiny{\textbf{\textsf{\phantom{p}U-NET\phantom{p}}}}\\
      \fbox{\includegraphics[width=1.0\linewidth]{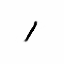}}
      \fbox{\includegraphics[width=1.0\linewidth]{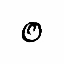}}
      \end{center}
   \end{minipage}
   \hspace{0.0025\linewidth}
   \begin{minipage}[t]{0.155\linewidth}
      \vspace{0pt}
      \begin{center}
      \tiny{\textbf{\textsf{Input}}}\\
      \fbox{\includegraphics[width=1.0\linewidth]{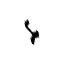}}
      \fbox{\includegraphics[width=1.0\linewidth]{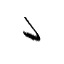}}
      \end{center}
   \end{minipage}
   \begin{minipage}[t]{0.155\linewidth}
      \vspace{0pt}
      \begin{center}
      \tiny{\textbf{\textsf{\phantom{p}True\phantom{p}}}}\\
      \fbox{\includegraphics[width=1.0\linewidth]{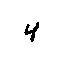}}
      \fbox{\includegraphics[width=1.0\linewidth]{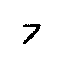}}
      \end{center}
   \end{minipage}
   \begin{minipage}[t]{0.155\linewidth}
      \vspace{0pt}
      \begin{center}
      \tiny{\textbf{\textsf{\phantom{p}U-NET\phantom{p}}}}\\
      \fbox{\includegraphics[width=1.0\linewidth]{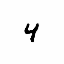}}
      \fbox{\includegraphics[width=1.0\linewidth]{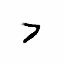}}
      \end{center}
   \end{minipage}\\
   \vspace{0.013\linewidth}
   \scriptsize{\textbf{\textsf{Affine Only}}}\\
   \begin{minipage}[t]{0.155\linewidth}
      \vspace{0pt}
      \begin{center}
      \tiny{\textbf{\textsf{Input}}}\\
      \fbox{\includegraphics[width=1.0\linewidth]{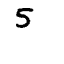}}
      \fbox{\includegraphics[width=1.0\linewidth]{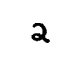}}
      \end{center}
   \end{minipage}
   \begin{minipage}[t]{0.155\linewidth}
      \vspace{0pt}
      \begin{center}
      \tiny{\textbf{\textsf{\phantom{p}True\phantom{p}}}}\\
      \fbox{\includegraphics[width=1.0\linewidth]{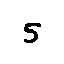}}
      \fbox{\includegraphics[width=1.0\linewidth]{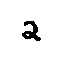}}
      \end{center}
   \end{minipage}
   \begin{minipage}[t]{0.155\linewidth}
      \vspace{0pt}
      \begin{center}
      \tiny{\textbf{\textsf{\phantom{p}U-NET\phantom{p}}}}\\
      \fbox{\includegraphics[width=1.0\linewidth]{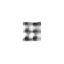}}
      \fbox{\includegraphics[width=1.0\linewidth]{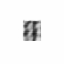}}
      \end{center}
   \end{minipage}
   \hspace{0.0025\linewidth}
   \begin{minipage}[t]{0.155\linewidth}
      \vspace{0pt}
      \begin{center}
      \tiny{\textbf{\textsf{Input}}}\\
      \fbox{\includegraphics[width=1.0\linewidth]{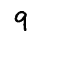}}
      \fbox{\includegraphics[width=1.0\linewidth]{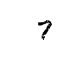}}
      \end{center}
   \end{minipage}
   \begin{minipage}[t]{0.155\linewidth}
      \vspace{0pt}
      \begin{center}
      \tiny{\textbf{\textsf{\phantom{p}True\phantom{p}}}}\\
      \fbox{\includegraphics[width=1.0\linewidth]{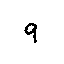}}
      \fbox{\includegraphics[width=1.0\linewidth]{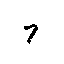}}
      \end{center}
   \end{minipage}
   \begin{minipage}[t]{0.155\linewidth}
      \vspace{0pt}
      \begin{center}
      \tiny{\textbf{\textsf{\phantom{p}U-NET\phantom{p}}}}\\
      \fbox{\includegraphics[width=1.0\linewidth]{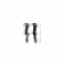}}
      \fbox{\includegraphics[width=1.0\linewidth]{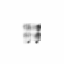}}
      \end{center}
   \end{minipage}\\
   \vspace{0.013\linewidth}
   \scriptsize{\textbf{\textsf{Translation Only}}}
   \end{center}
\end{minipage}

%% file: include/nmg/fig_sal_content.tex
\setlength\fboxsep{0pt}
\begin{minipage}[t]{0.32\linewidth}
   \vspace{0pt}
   \begin{center}
   \begin{minipage}[t]{0.19\linewidth}
      \vspace{0pt}
      \begin{center}
         \tiny{\textbf{\textsf{Input}}}\\
         \fbox{\includegraphics[width=1.0\linewidth]{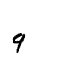}}
      \end{center}
   \end{minipage}
   \hfill
   \begin{minipage}[t]{0.79\linewidth}
      \vspace{0pt}
      \begin{center}
         \begin{minipage}[t]{0.94\linewidth}
            \vspace{0pt}
            \begin{center}
            \begin{minipage}[t]{0.32\linewidth}
               \vspace{0pt}
               \begin{center}
               \tiny{\textbf{\textsf{Output}}}\\
               \fbox{\includegraphics[width=1.0\linewidth]{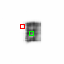}}
               \fbox{\includegraphics[width=1.0\linewidth]{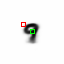}}
               \fbox{\includegraphics[width=1.0\linewidth]{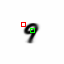}}
               \end{center}
            \end{minipage}
            \begin{minipage}[t]{0.32\linewidth}
               \vspace{0pt}
               \begin{center}
               \tiny{\textbf{\textsf{\phantom{p}Attention\phantom{p}}}}\\
               \fbox{\includegraphics[width=1.0\linewidth]{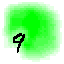}}
               \fbox{\includegraphics[width=1.0\linewidth]{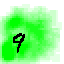}}
               \fbox{\includegraphics[width=1.0\linewidth]{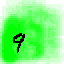}}
               \end{center}
            \end{minipage}
            \begin{minipage}[t]{0.32\linewidth}
               \vspace{0pt}
               \begin{center}
               \tiny{\textbf{\textsf{\phantom{p}Attention\phantom{p}}}}\\
               \fbox{\includegraphics[width=1.0\linewidth]{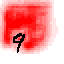}}
               \fbox{\includegraphics[width=1.0\linewidth]{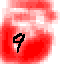}}
               \fbox{\includegraphics[width=1.0\linewidth]{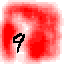}}
               \end{center}
            \end{minipage}
            \end{center}
         \end{minipage}
         \begin{minipage}[t]{0.04\linewidth}
            \vspace{0pt}
            \begin{rotate}{270}
               \tiny{\textbf{\textsf{
                  \hspace*{15pt}
                  U-NET
                  \hspace*{23.5pt}
                  PMG
                  \hspace*{22pt}
                  R-PMG
               }}}
            \end{rotate}
         \end{minipage}
      \end{center}
   \end{minipage}
   \end{center}
\end{minipage}
\hfill
\begin{minipage}[t]{0.32\linewidth}
   \vspace{0pt}
   \begin{center}
   \begin{minipage}[t]{0.19\linewidth}
      \vspace{0pt}
      \begin{center}
         \tiny{\textbf{\textsf{Input}}}\\
         \fbox{\includegraphics[width=1.0\linewidth]{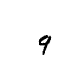}}
      \end{center}
   \end{minipage}
   \hfill
   \begin{minipage}[t]{0.79\linewidth}
      \vspace{0pt}
      \begin{center}
         \begin{minipage}[t]{0.94\linewidth}
            \vspace{0pt}
            \begin{center}
            \begin{minipage}[t]{0.32\linewidth}
               \vspace{0pt}
               \begin{center}
               \tiny{\textbf{\textsf{Output}}}\\
               \fbox{\includegraphics[width=1.0\linewidth]{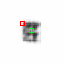}}
               \fbox{\includegraphics[width=1.0\linewidth]{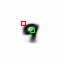}}
               \fbox{\includegraphics[width=1.0\linewidth]{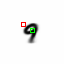}}
               \end{center}
            \end{minipage}
            \begin{minipage}[t]{0.32\linewidth}
               \vspace{0pt}
               \begin{center}
               \tiny{\textbf{\textsf{\phantom{p}Attention\phantom{p}}}}\\
               \fbox{\includegraphics[width=1.0\linewidth]{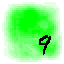}}
               \fbox{\includegraphics[width=1.0\linewidth]{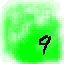}}
               \fbox{\includegraphics[width=1.0\linewidth]{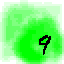}}
               \end{center}
            \end{minipage}
            \begin{minipage}[t]{0.32\linewidth}
               \vspace{0pt}
               \begin{center}
               \tiny{\textbf{\textsf{\phantom{p}Attention\phantom{p}}}}\\
               \fbox{\includegraphics[width=1.0\linewidth]{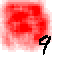}}
               \fbox{\includegraphics[width=1.0\linewidth]{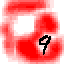}}
               \fbox{\includegraphics[width=1.0\linewidth]{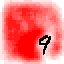}}
               \end{center}
            \end{minipage}
            \end{center}
         \end{minipage}
         \begin{minipage}[t]{0.04\linewidth}
            \vspace{0pt}
            \begin{rotate}{270}
               \tiny{\textbf{\textsf{
                  \hspace*{15pt}
                  U-NET
                  \hspace*{23.5pt}
                  PMG
                  \hspace*{22pt}
                  R-PMG
               }}}
            \end{rotate}
         \end{minipage}
      \end{center}
   \end{minipage}
   \end{center}
\end{minipage}
\hfill
\begin{minipage}[t]{0.32\linewidth}
   \vspace{0pt}
   \begin{center}
   \begin{minipage}[t]{0.19\linewidth}
      \vspace{0pt}
      \begin{center}
         \tiny{\textbf{\textsf{Input}}}\\
         \fbox{\includegraphics[width=1.0\linewidth]{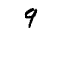}}
      \end{center}
   \end{minipage}
   \hfill
   \begin{minipage}[t]{0.79\linewidth}
      \vspace{0pt}
      \begin{center}
         \begin{minipage}[t]{0.94\linewidth}
            \vspace{0pt}
            \begin{center}
            \begin{minipage}[t]{0.32\linewidth}
               \vspace{0pt}
               \begin{center}
               \tiny{\textbf{\textsf{Output}}}\\
               \fbox{\includegraphics[width=1.0\linewidth]{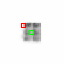}}
               \fbox{\includegraphics[width=1.0\linewidth]{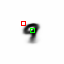}}
               \fbox{\includegraphics[width=1.0\linewidth]{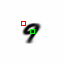}}
               \end{center}
            \end{minipage}
            \begin{minipage}[t]{0.32\linewidth}
               \vspace{0pt}
               \begin{center}
               \tiny{\textbf{\textsf{\phantom{p}Attention\phantom{p}}}}\\
               \fbox{\includegraphics[width=1.0\linewidth]{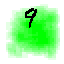}}
               \fbox{\includegraphics[width=1.0\linewidth]{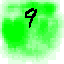}}
               \fbox{\includegraphics[width=1.0\linewidth]{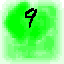}}
               \end{center}
            \end{minipage}
            \begin{minipage}[t]{0.32\linewidth}
               \vspace{0pt}
               \begin{center}
               \tiny{\textbf{\textsf{\phantom{p}Attention\phantom{p}}}}\\
               \fbox{\includegraphics[width=1.0\linewidth]{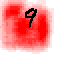}}
               \fbox{\includegraphics[width=1.0\linewidth]{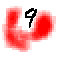}}
               \fbox{\includegraphics[width=1.0\linewidth]{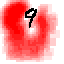}}
               \end{center}
            \end{minipage}
            \end{center}
         \end{minipage}
         \begin{minipage}[t]{0.04\linewidth}
            \vspace{0pt}
            \begin{rotate}{270}
               \tiny{\textbf{\textsf{
                  \hspace*{15pt}
                  U-NET
                  \hspace*{23.5pt}
                  PMG
                  \hspace*{22pt}
                  R-PMG
               }}}
            \end{rotate}
         \end{minipage}
      \end{center}
   \end{minipage}
   \end{center}
\end{minipage}\\
\vspace{0.01\linewidth}
\setlength\fboxsep{0pt}
\begin{minipage}[t]{0.32\linewidth}
   \vspace{0pt}
   \begin{center}
   \begin{minipage}[t]{0.19\linewidth}
      \vspace{0pt}
      \begin{center}
         \tiny{\textbf{\textsf{Input}}}\\
         \fbox{\includegraphics[width=1.0\linewidth]{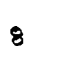}}
      \end{center}
   \end{minipage}
   \hfill
   \begin{minipage}[t]{0.79\linewidth}
      \vspace{0pt}
      \begin{center}
         \begin{minipage}[t]{0.94\linewidth}
            \vspace{0pt}
            \begin{center}
            \begin{minipage}[t]{0.32\linewidth}
               \vspace{0pt}
               \begin{center}
               \tiny{\textbf{\textsf{Output}}}\\
               \fbox{\includegraphics[width=1.0\linewidth]{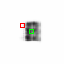}}
               \fbox{\includegraphics[width=1.0\linewidth]{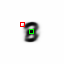}}
               \fbox{\includegraphics[width=1.0\linewidth]{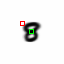}}
               \end{center}
            \end{minipage}
            \begin{minipage}[t]{0.32\linewidth}
               \vspace{0pt}
               \begin{center}
               \tiny{\textbf{\textsf{\phantom{p}Attention\phantom{p}}}}\\
               \fbox{\includegraphics[width=1.0\linewidth]{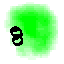}}
               \fbox{\includegraphics[width=1.0\linewidth]{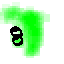}}
               \fbox{\includegraphics[width=1.0\linewidth]{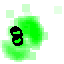}}
               \end{center}
            \end{minipage}
            \begin{minipage}[t]{0.32\linewidth}
               \vspace{0pt}
               \begin{center}
               \tiny{\textbf{\textsf{\phantom{p}Attention\phantom{p}}}}\\
               \fbox{\includegraphics[width=1.0\linewidth]{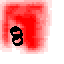}}
               \fbox{\includegraphics[width=1.0\linewidth]{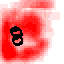}}
               \fbox{\includegraphics[width=1.0\linewidth]{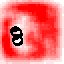}}
               \end{center}
            \end{minipage}
            \end{center}
         \end{minipage}
         \begin{minipage}[t]{0.04\linewidth}
            \vspace{0pt}
            \begin{rotate}{270}
               \tiny{\textbf{\textsf{
                  \hspace*{15pt}
                  U-NET
                  \hspace*{23.5pt}
                  PMG
                  \hspace*{22pt}
                  R-PMG
               }}}
            \end{rotate}
         \end{minipage}
      \end{center}
   \end{minipage}
   \end{center}
\end{minipage}
\hfill
\begin{minipage}[t]{0.32\linewidth}
   \vspace{0pt}
   \begin{center}
   \begin{minipage}[t]{0.19\linewidth}
      \vspace{0pt}
      \begin{center}
         \tiny{\textbf{\textsf{Input}}}\\
         \fbox{\includegraphics[width=1.0\linewidth]{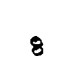}}
      \end{center}
   \end{minipage}
   \hfill
   \begin{minipage}[t]{0.79\linewidth}
      \vspace{0pt}
      \begin{center}
         \begin{minipage}[t]{0.94\linewidth}
            \vspace{0pt}
            \begin{center}
            \begin{minipage}[t]{0.32\linewidth}
               \vspace{0pt}
               \begin{center}
               \tiny{\textbf{\textsf{Output}}}\\
               \fbox{\includegraphics[width=1.0\linewidth]{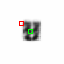}}
               \fbox{\includegraphics[width=1.0\linewidth]{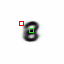}}
               \fbox{\includegraphics[width=1.0\linewidth]{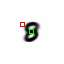}}
               \end{center}
            \end{minipage}
            \begin{minipage}[t]{0.32\linewidth}
               \vspace{0pt}
               \begin{center}
               \tiny{\textbf{\textsf{\phantom{p}Attention\phantom{p}}}}\\
               \fbox{\includegraphics[width=1.0\linewidth]{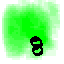}}
               \fbox{\includegraphics[width=1.0\linewidth]{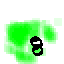}}
               \fbox{\includegraphics[width=1.0\linewidth]{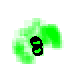}}
               \end{center}
            \end{minipage}
            \begin{minipage}[t]{0.32\linewidth}
               \vspace{0pt}
               \begin{center}
               \tiny{\textbf{\textsf{\phantom{p}Attention\phantom{p}}}}\\
               \fbox{\includegraphics[width=1.0\linewidth]{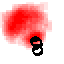}}
               \fbox{\includegraphics[width=1.0\linewidth]{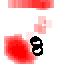}}
               \fbox{\includegraphics[width=1.0\linewidth]{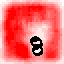}}
               \end{center}
            \end{minipage}
            \end{center}
         \end{minipage}
         \begin{minipage}[t]{0.04\linewidth}
            \vspace{0pt}
            \begin{rotate}{270}
               \tiny{\textbf{\textsf{
                  \hspace*{15pt}
                  U-NET
                  \hspace*{23.5pt}
                  PMG
                  \hspace*{22pt}
                  R-PMG
               }}}
            \end{rotate}
         \end{minipage}
      \end{center}
   \end{minipage}
   \end{center}
\end{minipage}
\hfill
\begin{minipage}[t]{0.32\linewidth}
   \vspace{0pt}
   \begin{center}
   \begin{minipage}[t]{0.19\linewidth}
      \vspace{0pt}
      \begin{center}
         \tiny{\textbf{\textsf{Input}}}\\
         \fbox{\includegraphics[width=1.0\linewidth]{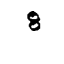}}
      \end{center}
   \end{minipage}
   \hfill
   \begin{minipage}[t]{0.79\linewidth}
      \vspace{0pt}
      \begin{center}
         \begin{minipage}[t]{0.94\linewidth}
            \vspace{0pt}
            \begin{center}
            \begin{minipage}[t]{0.32\linewidth}
               \vspace{0pt}
               \begin{center}
               \tiny{\textbf{\textsf{Output}}}\\
               \fbox{\includegraphics[width=1.0\linewidth]{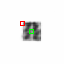}}
               \fbox{\includegraphics[width=1.0\linewidth]{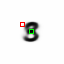}}
               \fbox{\includegraphics[width=1.0\linewidth]{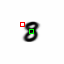}}
               \end{center}
            \end{minipage}
            \begin{minipage}[t]{0.32\linewidth}
               \vspace{0pt}
               \begin{center}
               \tiny{\textbf{\textsf{\phantom{p}Attention\phantom{p}}}}\\
               \fbox{\includegraphics[width=1.0\linewidth]{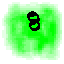}}
               \fbox{\includegraphics[width=1.0\linewidth]{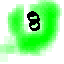}}
               \fbox{\includegraphics[width=1.0\linewidth]{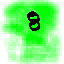}}
               \end{center}
            \end{minipage}
            \begin{minipage}[t]{0.32\linewidth}
               \vspace{0pt}
               \begin{center}
               \tiny{\textbf{\textsf{\phantom{p}Attention\phantom{p}}}}\\
               \fbox{\includegraphics[width=1.0\linewidth]{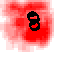}}
               \fbox{\includegraphics[width=1.0\linewidth]{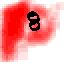}}
               \fbox{\includegraphics[width=1.0\linewidth]{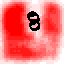}}
               \end{center}
            \end{minipage}
            \end{center}
         \end{minipage}
         \begin{minipage}[t]{0.04\linewidth}
            \vspace{0pt}
            \begin{rotate}{270}
               \tiny{\textbf{\textsf{
                  \hspace*{15pt}
                  U-NET
                  \hspace*{23.5pt}
                  PMG
                  \hspace*{22pt}
                  R-PMG
               }}}
            \end{rotate}
         \end{minipage}
      \end{center}
   \end{minipage}
   \end{center}
\end{minipage}

%% file: include/nmg/fig_sal_pipeline.tex
\setlength\fboxsep{0pt}
\begin{minipage}[t]{1.0\linewidth}
   \vspace{0pt}
   \begin{center}
   \begin{minipage}[t]{1.0\linewidth}
      \vspace{0pt}
      \begin{center}
      \begin{minipage}[t]{0.32\linewidth}
         \vspace{0pt}
         \begin{center}
         \scriptsize{\textbf{\textsf{Input}}}\\
         \fbox{\includegraphics[width=1.0\linewidth]{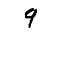}}
         \end{center}
      \end{minipage}
      \begin{minipage}[t]{0.32\linewidth}
         \vspace{0pt}
         \begin{center}
         \scriptsize{\textbf{\textsf{Output (R-PMG)}}}\\
         \fbox{\includegraphics[width=1.0\linewidth]{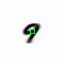}}
         \end{center}
      \end{minipage}
      \hfill
      \begin{minipage}[t]{0.32\linewidth}
         \vspace{0pt}
         \begin{center}
         \scriptsize{\textbf{\textsf{Attention Map}}}\\
         \fbox{\includegraphics[width=1.0\linewidth]{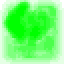}}
         \end{center}
      \end{minipage}
      \end{center}
   \end{minipage}\\
   \vspace{0.02\linewidth}
   \begin{minipage}[t]{1.0\linewidth}
      \vspace{0pt}
      \begin{center}
      \begin{minipage}[t]{0.2425\linewidth}
         \vspace{0pt}
         \begin{center}
         \tiny{\textbf{\textsf{Input}}}\\
         \fbox{\includegraphics[width=1.0\linewidth]{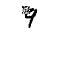}}
         \fbox{\includegraphics[width=1.0\linewidth]{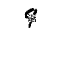}}
         \fbox{\includegraphics[width=1.0\linewidth]{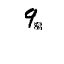}}
         \end{center}
      \end{minipage}
      \hfill
      \begin{minipage}[t]{0.2425\linewidth}
         \vspace{0pt}
         \begin{center}
         \tiny{\textbf{\textsf{Output}}}\\
         \fbox{\includegraphics[width=1.0\linewidth]{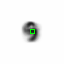}}
         \fbox{\includegraphics[width=1.0\linewidth]{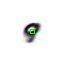}}
         \fbox{\includegraphics[width=1.0\linewidth]{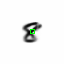}}
         \end{center}
      \end{minipage}
      \hfill
      \begin{minipage}[t]{0.2425\linewidth}
         \vspace{0pt}
         \begin{center}
         \tiny{\textbf{\textsf{Difference in Output}}}\\
         \fbox{\includegraphics[width=1.0\linewidth]{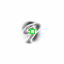}}
         \fbox{\includegraphics[width=1.0\linewidth]{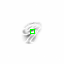}}
         \fbox{\includegraphics[width=1.0\linewidth]{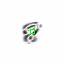}}
         \end{center}
      \end{minipage}
      \hfill
      \begin{minipage}[t]{0.2425\linewidth}
         \vspace{0pt}
         \begin{center}
         \tiny{\textbf{\textsf{Map Contribution}}}\\
         \fbox{\includegraphics[width=1.0\linewidth]{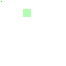}}
         \fbox{\includegraphics[width=1.0\linewidth]{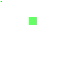}}
         \fbox{\includegraphics[width=1.0\linewidth]{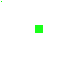}}
         \end{center}
      \end{minipage}
      \end{center}
   \end{minipage}
   \end{center}
\end{minipage}

%% file: include/nmg/sec5_conclusion.tex
\section{Conclusion}
\label{sec:conclusion}

Our proposed multigrid extension to CNNs yields improved accuracy on
classification and semantic segmentation tasks.  Progressive multigrid
variants open a new pathway towards optimizing CNNs for efficiency.
Multigrid appears unique in extending the range of tasks CNNs can accomplish,
by integrating into the network structure the capacity to learn routing and
attentional mechanisms.  These new abilities suggest that multigrid could
replace some ad-hoc designs in the current zoo of CNN architectures.

On a speculative note, multigrid neural networks might also have broader
implications in neuroscience.  Feedforward computation on a sequence of
multigrid pyramids looks similar to combined bottom-up/top-down processing
across a single larger structure if neurons are embedded in some computational
substrate according to their spatial grid, rather than their depth in the
processing chain.